\documentclass[10pt,twocolumn,letterpaper]{article}

\usepackage{cvpr}
\usepackage{times}
\usepackage{epsfig}
\usepackage{graphicx}
\usepackage{amsmath}
\usepackage{amssymb}
\usepackage{multirow}
\usepackage{subfigure}
\usepackage{algorithm}
\usepackage{algorithmic}

\usepackage[pagebackref=true,breaklinks=true,letterpaper=true,colorlinks,bookmarks=false]{hyperref}

\cvprfinalcopy 

\def\cvprPaperID{1004} 

\ifcvprfinal\fi
\begin{document}

\title{Collaborative Quantization for Cross-Modal Similarity Search}

\author{Ting Zhang\textsuperscript{1}\thanks{This work was done when Ting Zhang was an intern at MSR.}\quad
Jingdong Wang\textsuperscript{2}\\
\textsuperscript{1}University of Science and Technology of China, China\quad
\textsuperscript{2}Microsoft Research, China\\
\textsuperscript{1}{\tt\small zting@mail.ustc.edu.cn}\quad
\textsuperscript{2}{\tt\small jingdw@microsoft.com}\\
}

\maketitle
\thispagestyle{empty}

\begin{abstract}
Cross-modal similarity search
is a problem about designing a search system
supporting querying across content modalities,
e.g., using an image to search for texts
or using a text to search for images.
This paper presents a compact coding solution
for efficient search,
with a focus on the quantization approach
which has already shown the superior performance
over the hashing solutions
in the single-modal similarity search.
We propose a cross-modal quantization approach,
which is among the early attempts
to introduce quantization into cross-modal search.
The major contribution lies in jointly
learning the quantizers for both modalities
through aligning the quantized representations
for each pair of image and text belonging to
a document.
In addition,
our approach simultaneously
learns the common space
for both modalities
in which quantization is conducted
to enable efficient and effective search
using the Euclidean distance computed
in the common space
with fast distance table lookup.
Experimental results compared with several competitive
algorithms over three benchmark datasets
demonstrate that the proposed approach achieves the state-of-the-art performance.

\end{abstract}

\section{Introduction}

Similarity search has been a fundamental problem
in information retrieval and multimedia search.
Classical approaches, however, are designed to address
the single-modal search problem~\cite{WangSSJ14,WangL12, WangWZGLG13, WangWJLZZH13, MujaL09, MujaL14}, where, for instance,
the text query is used to search in a text database,
or the image query is used to search in an image database.
In this paper,
we deal with the cross-modal similarity search problem,
which is an important problem emerged in multimedia information retrieval,
for example, using a text query to retrieve images
or using an image query to retrieve texts.

We study the compact coding solutions to cross-modal similarity search,
in particular focusing on
a common real-world scenario, image and text modalities.
Compact coding is an approach
of converting the database items into short codes
on which similarity search can be efficiently conducted.
It has been widely studied
in single-modal similarity search
with typical solutions including hashing~\cite{DatarIIM04,NorouziF11,WangKC12} and quantization~\cite{GongLGP13,JegouDS11,0002F13,ZhangDW14, WangWSXSL15, ZhangQTW15},
while relatively unexplored in cross-modal search
except a few hashing approaches~\cite{BronsteinBMP10,KumarU11,MasciBBS14,ZhouDG14}.
We are interested in the quantization approach
that represents each point by a short code
formed by the index of the nearest center,
as quantization has shown more
powerful representation ability than hashing
in single-modal search.

Rather than performing the quantization directly
in the original feature space,
we learn a common space for both modalities
with the goal that the pair of image and text lie
in the learnt common space closely.
Learning such a common space
is important and useful
for the subsequent quantization
whose similarity is computed based on the Euclidean distance.
Similar observation  has also been made in some hashing techniques~\cite{RasiwasiaMV07,RasiwasiaPCDLLV10,ZhouDG14}
that
apply the sign function on the learnt common space.

In this paper, we propose a novel approach for cross-modal similarity search,
called collaborative quantization,
that conducts the quantization simultaneously
for both modalities in the common space,
to which the database items of both modalities are
mapped through matrix factorization.
The quantization and the common space mapping
are jointly optimized
for both modalities
under the objective that
the quantized approximations of the descriptors
of an image and a text
forming a pair  in the search database
are well aligned.
Our approach is one of the early attempts
to introduce quantization into cross-modal
similarity search
offering the superior search performance.
Experimental results on several standard datasets
show that
our approach outperforms existing cross-modal hashing and quantization algorithms.

\section{Related work}
\begin{table*} \footnotesize
\centering
\caption{A brief categorization of the compact coding algorithms for cross-modal similarity search.
The multi-modal relations are roughly divided
into four categories:
intra-modality (image vs. image and text vs. text),
inter-modality (image vs. text),
intra-document (correspondence of an image and a text forming a document,
a special kind of inter-modality),
and inter-document (document vs. document).
Unified codes denote that the codes for an image and a text belonging to a document are the same,
and separate codes denote that the codes are different.}
\label{tab:categorization}
\begin{tabular}{ |l||cccc|cc|cc| }
\hline
\multirow{2}{*}{Methods}& \multicolumn{4}{c|}{Multi-modal data relations}  & \multicolumn{2}{c|}{Codes}  & \multicolumn{2}{c|}{Coding methods}\\
\cline{2-9}
 & Intra-modality & Inter-modality & Intra-document & Inter-document   & Unified & Separate & Hash & Quantization\\
 \hline
 \hline
 CMSSH~\cite{BronsteinBMP10} & & $\triangle$ & & & & $\triangle$ & $\triangle$ & \\
 \hline
 SCM~\cite{ZhangL14} & & $\triangle$ & & &  & $\triangle$ & $\triangle$ & \\
  \hline
  CRH~\cite{ZhenY12b} &  & $\triangle$ & &  & & $\triangle$ & $\triangle$ & \\
 \hline
 MMNN~\cite{MasciBBS14} & $\triangle$ & $\triangle$ & &  & & $\triangle$ & $\triangle$ & \\
 \hline
 $\textrm{SM}^2\textrm{H}$~\cite{WuYYTZZ14} & $\triangle$ & $\triangle$ & &  & & $\triangle$ & $\triangle$ & \\
 \hline
  MLBE~\cite{ZhenY12a} & $\triangle$ & $\triangle$ & & & & $\triangle$ & $\triangle$ & \\
 \hline
 IMH~\cite{SongYYHS13} & $\triangle$ &  & $\triangle$ & & & $\triangle$ & $\triangle$ & \\
 \hline
 CVH~\cite{KumarU11} &  &  & $\triangle$ & $\triangle$ &  & $\triangle$ & $\triangle$ & \\
 \hline
 MVSH~\cite{KimKC12} & &  & $\triangle$ & $\triangle$ &  & $\triangle$ & $\triangle$ & \\
  \hline
 SPH~\cite{LinDH015} & &  &  $\triangle$ & $\triangle$ & $\triangle$ &  & $\triangle$ & \\
 \hline
 LSSH~\cite{ZhouDG14} & &  & $\triangle$ &   & $\triangle$ &  & $\triangle$ & \\
 \hline
 CMFH~\cite{DingGZ14}  & &  & $\triangle$ &   & $\triangle$ &  & $\triangle$ & \\
 \hline
 STMH~\cite{WangGWH15} & &  & $\triangle$ &   & $\triangle$ &  & $\triangle$ & \\
 \hline
 QCH~\cite{WuYZWW15} & & $\triangle$ &  &   &  & $\triangle$ & $\triangle$ & \\
  \hline
 CCQ~\cite{LongWY15} & &  & $\triangle$ &   & $\triangle$ &  &  & $\triangle$ \\
   \hline
  Our approach & &  & $\triangle$ &   &  & $\triangle$ &  & $\triangle$ \\
\hline
\end{tabular}
\vspace{-0.5cm}
\end{table*}
There are two categories of compact coding approaches
 for cross-modal similarity search:
cross-modal hashing and cross-modal quantization.

Cross-modal hashing often maps multi-modal data
into a common Hamming space
so that the hash codes
of different modalities are directly comparable
using the Hamming distance.
After mapping, each document may have just one unified hash code,
in which all the modalities of the document are mapped,
or may have two separate hash codes, each corresponding to a modality.
The main research problem in cross-modal hashing,
besides hash function design that is also studied in single-modal search,
is how to exploit and build the relations between the modalities.
In general, the relations of multi-modal data, besides the intra-modality relation in the single modality (image vs. image and text vs. text) and the inter-modality relation across the modalities (image vs. text),
also include
intra-document (the correspondence of an image and a text forming a document, which is a special kind of inter-modality)
and inter-document (document vs. document).
A brief categorization is shown in Table~\ref{tab:categorization}.

The early approach, data fusion hashing~\cite{BronsteinBMP10},
is a pairwise cross-modal similarity sensitive approach,
which aligns the similarities (defined as inner product) in the Hamming space
across the modalities,
with the given \emph{inter-modality}
similar and dissimilar relations
using the maximizing similarity-agreement criterion.
An alternative formulation using the minimizing similarity-difference criterion
is introduced in~\cite{ZhangL14}.
Co-regularized hashing~\cite{ZhenY12b}
uses a smoothly clipped inverted squared deviation function
to connect the inter-modality relation
with the similarity over the projections
that form the hashing codes.
Similar regularization techniques are adopted for multi-modal hashing in~\cite{MoranL15}.
In addition to the inter-modality similarities,
several other hashing techniques,
such as
multimodal similarity-preserving hashing~\cite{MasciBBS14},
sparse hashing approach~\cite{WuYYTZZ14},
a probabilistic model for hashing~\cite{ZhenY12a},
also explore and utilize the \emph{intra-modality} relation to
learn the hash codes for each modality.


Cross-view hashing~\cite{KumarU11}
defines the distance between documents
in the Hamming space
by considering the hash codes of all the modalities,
and aligns it with
the given \emph{inter-document} similarity.
Multi-view spectral hashing~\cite{KimKC12}
adopts a similar formulation
but with a different optimization algorithm.
These methods usually also involve
the intra-document relation in an implicit way
by considering the multi-modal document as an integrated whole object.
There are other hashing methods
exploring the inter-document relation
about multi-modal representation , but not for cross-modal similarity search,
such as composite hashing~\cite{ZhangWS11} and
effective multiple feature hashing~\cite{SongYHSL13}.




The \emph{intra-document} relation is often used
to learn a unified hash code,
into which a hash function is learnt for each modality
to map the feature.
For example,
Latent semantic sparse hashing~\cite{ZhouDG14}
applies the sign function
on the joint space projected from the latent semantic representation
learnt for each modality.
Collective matrix factorization hashing~\cite{DingGZ14}
finds the common (same) representation for an image-text pair
via collective matrix factorization,
and obtains the hash codes directly
using the sign function on the common representation.
Other methods exploring the intra-document relation
include
semantic topic multimodal hashing~\cite{WangGWH15},
semantics-preserving multi-view hashing~\cite{LinDH015},
inter-media hashing~\cite{ZhangWS11}
and its accelerated version~\cite{ZhuHSZ13},
and so on.
Meanwhile, several attempts~\cite{WangOYZZ14,SrivastavaS14}
have been made
based on the neural network
which can also be combined with
our approach to learn the common space.




Recently, a few techniques based on quantization are developed
for cross-modal search.
Quantized correlation hashing~\cite{WuYZWW15}
combines the hash function learning with the quantization
by
minimizing the inter-modality similarity disagreement
as well as the binary quantization  simultaneously.
Compositional correlation quantization~\cite{LongWY15}
projects the multi-modal data into a common space,
and then obtains a unified quantization representation for each document.
Our approach, also exploring the intra-document relation, belongs to this cross-modal quantization category
and achieves the state-of-the-art performance.

\section{Formulation}
We study the similarity search problem
over a database $\mathcal{Z}$ of documents
with two modalities: image and text.
Each document is a pair of image and text,
$\mathcal{Z} = \{(\mathbf{x}_n, \mathbf{y}_n)\}_{n=1}^N$,
where $\mathbf{x}_n \in \mathbb{R}^{D_I}$ is a $D_I$-dimensional feature vector
describing an image,
and $\mathbf{y}_n \in \mathbb{R}^{D_T}$ is a $D_T$-dimensional feature vector
describing a text.
Splitting the database $\mathcal{Z}$
yields two databases each formed by images and texts separately, i.e.,
$\mathcal{X} = \{\mathbf{x}_1, \mathbf{x}_2, \cdots, \mathbf{x}_N\}$
and $\mathcal{Y} = \{\mathbf{y}_1, \mathbf{y}_2, \cdots, \mathbf{y}_N\}$.
Given a image (text) query $\mathbf{x}_q$ ($\mathbf{y}_q$),
the goal of cross-modality similarity search
is to retrieve the closest match in the text (image) database:
$\arg\max_{\mathbf{y} \in \mathcal{Y}} \operatorname{sim}(\mathbf{x}_q, \mathbf{y})$
($\arg\max_{\mathbf{x} \in \mathcal{X}} \operatorname{sim}(\mathbf{y}_q, \mathbf{x})$).

Rather than directly quantizing the feature vectors $\mathbf{x}$ and $\mathbf{y}$
to $\bar{\mathbf{x}}$ and $\bar{\mathbf{y}}$,
which requires a further non-trivial scheme to learn the similarity
for vectors $\bar{\mathbf{x}}$ and $\bar{\mathbf{y}}$
with different dimensions,
we are interested in finding
the common space for both image and text,
and jointly quantizing the image and text descriptors
in the common space,
so that
the Euclidean distance
which is widely-used in single-modal similarity search,
can also be used
for the cross-modal similarity evaluation.

\noindent\textbf{Collaborative quantization.}
Suppose the images and the texts
in the $D$-dimensional common space
are represented as
$\mathbf{X}' = [\mathbf{x}'_1, \mathbf{x}'_2, \cdots, \mathbf{x}'_N]$
and $\mathbf{Y}' = [\mathbf{y}'_1, \mathbf{y}'_2, \cdots, \mathbf{y}'_N]$.
For each modality,
we propose to adopt composite quantization~\cite{ZhangDW14}
to quantize the vectors in the common space.
Composite quantization aims to
approximate the images $\mathbf{X}'$ as
$\mathbf{X}' \approx \bar{\mathbf{X}} = \mathbf{C}\mathbf{P}$
by minimizing
\begin{align}
\|\mathbf{X}' - \mathbf{C}\mathbf{P}\|_F^2.
\end{align}
Here, $\mathbf{C} = [\mathbf{C}_1, \mathbf{C}_2, \cdots, \mathbf{C}_M]$ corresponds
to the $M$ dictionaries,
$\mathbf{C}_m = [\mathbf{c}_{m1},\mathbf{c}_{m2}, \cdots ,\mathbf{c}_{mK}]$ corresponds to the $m$th dictionary of size $K$
and each column is a dictionary element.
$\mathbf{P} = [\mathbf{p}_1, \mathbf{p}_2,\cdots,\mathbf{p}_N]$
with
$\mathbf{p}_n = [\mathbf{p}_{n1}^T,\mathbf{p}_{n2}^T,\cdots,\mathbf{p}_{nm}^T]^T$,
and
$\mathbf{p}_{nm}$ is a $K$-dimensional binary ($0$,$1$) vector
with only $1$-valued entry
indicating that the corresponding element in the $m$th dictionary
is selected to compose $\mathbf{x}'_n$.
The texts $\mathbf{Y}'$ in the common space
are approximated
as
$\mathbf{Y}' \approx \bar{\mathbf{Y}} =\mathbf{D}\mathbf{Q}$,
and the meaning of the symbols is similar to
that in the images.

Besides the quantization quality,
we explore the intra-document correlation between images and texts for the quantization:
the image and the text forming a document
are close after quantization,
which is the bridge to connect images and texts
for cross-modal search.
We adopt the following simple formulation
that
minimizes the distance
between the image and the corresponding text,
\begin{align}
\|\mathbf{C}\mathbf{P} - \mathbf{D}\mathbf{Q}\|_F^2.
\end{align}

The overall collaborative quantization formulation is given as follows,
\begin{align}
\label{eqn:quantization}
& \mathcal{Q} (\mathbf{C}, \mathbf{P}; \mathbf{D}, \mathbf{Q}) =\\
& \|\mathbf{X}' - \mathbf{C}\mathbf{P}\|^2_F + \|\mathbf{Y}' - \mathbf{D}\mathbf{Q}\|^2_F
+ \gamma\|\mathbf{C}\mathbf{P} - \mathbf{D}\mathbf{Q}\|^2_F, \nonumber
\end{align}
where $\gamma$ is a trade-off variable
to balance the quantization quality
and the correlation degree.

\noindent\textbf{Common space mapping.}
The common space mapping problem
aims to map the data in different modalities
into the same space so that the representations in cross-modalities
are comparable.
In our problem, we want to map the $N$ $D_I$-dimensional image data $\mathbf{X}$
and the $N$ $D_T$-dimensional text data $\mathbf{Y}$
to the same $D$-dimensional data: $\mathbf{X}'$ and $\mathbf{Y}'$.

We choose the matrix-decomposition solution as in~\cite{ZhouDG14}:
the image data $\mathbf{X}$ is approximated using sparse coding
as a product of two matrices
$\mathbf{B}\mathbf{S}$,
and the sparse code $\mathbf{S}$ is shown to be a good representation
of the raw feature $\mathbf{X}$;
the text data $\mathbf{Y}$ is also decomposed into two matrices,
$\mathbf{U}$ and $\mathbf{Y}'$,
where $\mathbf{Y}'$ is the low-dimensional representation;
In addition,
a transformation matrix $\mathbf{R}$ is introduced to align the image sparse code
$\mathbf{S}$
with the text code $\mathbf{Y}'$
by minimizing
$\|\mathbf{Y}' - \mathbf{R}\mathbf{S}\|_F^2$,
and the image
in the common space
is represented as $\mathbf{X}' = \mathbf{R}\mathbf{S}$.
The objective function for common space mapping
is written as follows,
\begin{align}
\label{eqn:commonspace}
& \mathcal{M} (\mathbf{B}, \mathbf{S}; \mathbf{U}, \mathbf{Y}'; \mathbf{R}) =\\
&\| \mathbf{X} - \mathbf{B}\mathbf{S} \|^2_F + \rho |\mathbf{S}|_{11}
+ \eta \|\mathbf{Y} - \mathbf{U} \mathbf{Y}'\|^2_F
+ \lambda \|\mathbf{Y}' - \mathbf{R}\mathbf{S}\|^2_F. \nonumber
\end{align}
Here $|\mathbf{S}|_{11} = \sum_{i=1}^N\|\mathbf{S}_{\cdot i}\|_1$ is the sparse term, and $\rho$ determines the sparsity degree;
$\eta$ is used to balance the scales of image and text representations;
$\lambda$ is a trade-off parameter
to control the approximation degree in each modality
and the alignment degree for the pair of image and text.

\noindent\textbf{Overall objective function.}
In summary, the overall formulation of the proposed cross-modal quantization
is,
\begin{align}
\label{eqn:overallfunction}
\min ~&~ \mathcal{F} (\boldsymbol{\theta}_q, \boldsymbol{\theta}_m) = \mathcal{Q} (\mathbf{C}, \mathbf{P}; \mathbf{D}, \mathbf{Q}) +
\mathcal{M} (\mathbf{B}, \mathbf{S}; \mathbf{U}, \mathbf{Y}'; \mathbf{R}) \nonumber \\
\operatorname{s.t.} ~&~ \|\mathbf{B}_{\cdot i}\|_2^2 \leqslant 1, \|\mathbf{U}_{\cdot i}\|_2^2 \leqslant 1,
\|\mathbf{R}_{\cdot i}\|_2^2 \leqslant 1, \\
~&~ \sum\nolimits_{i=1}^M \sum\nolimits_{j=1, j\neq i}^M  \mathbf{p}_{ni}^T \mathbf{C}_{i}^T\mathbf{C}_{j}\mathbf{p}_{nj} = \epsilon_1,  \label{eqn:constraint1}\\
  ~&~ \sum\nolimits_{i=1}^M \sum\nolimits_{j=1, j\neq i}^M  \mathbf{q}_{ni}^T \mathbf{D}_{i}^T\mathbf{D}_{j}\mathbf{q}_{nj} = \epsilon_2, \label{eqn:constraint2}
\end{align}
where $\boldsymbol{\theta}_q$ and $\boldsymbol{\theta}_m$
represent the parameters in quantization and mapping, i.e.,
$(\mathbf{C}, \mathbf{P}; \mathbf{D}, \mathbf{Q})$
and $(\mathbf{B}, \mathbf{S}; \mathbf{U}, \mathbf{Y}'; \mathbf{R})$ respectively.
The constraints in Equation~\ref{eqn:constraint1} and Equation~\ref{eqn:constraint2}
are introduced for fast distance computation as
in composite quantization~\cite{ZhangDW14},
and
more details about the search process are presented in
Section~\ref{sec:searchprocess}.

\section{Optimization}
We optimize the Problem~\ref{eqn:overallfunction}
by alternatively solving two sub-problems:
common space mapping with the quantization parameters fixed:
$\min \mathcal{F}(\boldsymbol{\theta}_m| \boldsymbol{\theta}_q) =
\mathcal{M} (\boldsymbol{\theta}_m) + \|\mathbf{X}' - \mathbf{C}\mathbf{P}\|^2_F + \|\mathbf{Y}' - \mathbf{D}\mathbf{Q}\|^2_F$,
and collaborative quantization with the
mapping parameters fixed:
$\min \mathcal{F}(\boldsymbol{\theta}_q| \boldsymbol{\theta}_m)
= \min \mathcal{Q}(\boldsymbol{\theta}_q)$.
Each of the two sub-problems is solved again
by a standard iteratively alternative algorithm.

\subsection{Common space mapping}
The objective function of the common space mapping with the quantization parameters fixed is,
\begin{align}
\min_{\boldsymbol{\theta}_m} ~&
\mathcal{M} (\boldsymbol{\theta}_m) + \|\mathbf{X}' - \mathbf{C}\mathbf{P}\|^2_F + \|\mathbf{Y}' - \mathbf{D}\mathbf{Q}\|^2_F \\
\operatorname{s.t.} ~&~ \|\mathbf{B}_{\cdot i}\|_2^2 \leqslant 1, \|\mathbf{U}_{\cdot i}\|_2^2 \leqslant 1,
\|\mathbf{R}_{\cdot i}\|_2^2 \leqslant 1.
\end{align}
The iteration details are given below.

\noindent \textbf{Update $\mathbf{Y}'$.}
The objective function with respect to $\mathbf{Y}'$ is
an unconstrained quadratic optimization problem,
and is
solved by the following closed-form solution,
\begin{align}
\mathbf{Y}'^{*} = (\eta \mathbf{U}^T\mathbf{U} + (\lambda+1)\mathbf{I})^{-1} (\mathbf{D}\mathbf{Q} + \eta\mathbf{U}^T \mathbf{Y} + \lambda \mathbf{R}\mathbf{S}), \nonumber
\end{align}
where $\mathbf{I}$ is the identity matrix.

\noindent \textbf{Update $\mathbf{S}$.}
The objective function with respect to $\mathbf{S}$
can be transformed to,
\begin{align} \small
\min_{\mathbf{S}}~& \|\left[
   \begin{array}{c}
\sqrt {\frac {1} {\lambda+1}}\mathbf{X}\\
\frac 1 {\lambda+1} (\mathbf{CP} + \lambda \mathbf{A})
\end{array}
\right] - \left[
\begin{array}{c}
\sqrt {\frac {1} {\lambda+1}} \mathbf{B}\\
\mathbf{R}
\end{array}
\right]
\mathbf{S}
      \|_F^2  \nonumber \\
       &+\frac {\rho} {\lambda + 1}  |\mathbf{S}|_{11},
\end{align}
which is solved using the sparse learning with efficient projections
package\footnote{http://parnec.nuaa.edu.cn/jliu/largeScaleSparseLearning.htm}.

\noindent \textbf{Update $\mathbf{U,B,R} $.}
The algorithms for updating $\mathbf{U,B,R}$
are the same, as we can see from the following formulas,
\begin{align}
\min_{\mathbf{U}}~& \|\mathbf{Y} - \mathbf{U} \mathbf{Y}'\|^2_F,~~
\operatorname{s.t.} ~ \|\mathbf{U}_{\cdot i}\|_2^2 \leqslant 1, \\
\min_{\mathbf{B}}~& \|\mathbf{X} - \mathbf{B} \mathbf{S}\|^2_F,~~
\operatorname{s.t.} ~ \|\mathbf{B}_{\cdot i}\|_2^2 \leqslant 1, \\
\min_{\mathbf{R}}~& \|\frac 1 {\lambda+1} (\mathbf{CP}+ \lambda \mathbf{Y}') - \mathbf{RS} \|^2_F,~~
\operatorname{s.t.} ~ \|\mathbf{R}_{\cdot i}\|_2^2 \leqslant 1. \nonumber
\end{align}
All of the above three learning problems are minimizing
the quadratically constrained least square problem,
which has been well studied in numerical optimization field
and can be readily solved using the primal-dual conjugate gradient method.

\subsection{Collaborative quantization}
The second sub-problem is
transformed to an unconstrained formulation
by adding the equality constraints as a penalty regularization
with a penalty parameter $\mu$,
\begin{align}
&
\Psi = \mathcal{Q}(\boldsymbol{\theta}_q)
+   \mu \sum\nolimits_{n=1}^{N}(\sum\nolimits_{i \neq j}^M \mathbf{p}_{ni}^T \mathbf{C}_{i}^T\mathbf{C}_{j}\mathbf{p}_{nj} - \epsilon_1)^2 \nonumber \\
 & + \mu \sum\nolimits_{n=1}^{N}(\sum\nolimits_{i \neq j}^M \mathbf{q}_{ni}^T \mathbf{D}_{i}^T\mathbf{D}_{j}\mathbf{q}_{nj} - \epsilon_2)^2,
\end{align}
which is solved by alternatively updating each variable with others fixed.

\noindent \textbf{Update $\mathbf{C}$ ($\mathbf{D}$).}
The optimization procedures for $\mathbf{C}$ and $\mathbf{D}$
are essentially the same, so we only show how to optimize $\mathbf{C}$.
We adopt the
L-BFGS\footnote{http://www.ece.northwestern.edu/˜nocedal/lbfgs.html}
algorithm, one of the most frequently-used quasi-Newton methods,
to solve the unconstrained non-linear
problem with respect to $\mathbf{C}$.
The derivative of the objective function is
$[\frac {\partial \Psi} {\mathbf{C}_1}, \cdots, \frac {\partial \Psi} {\mathbf{C}_M}]$,
\begin{align} \small
&\frac {\partial \Psi} {\partial \mathbf{C}_m} = 2( (\gamma+1)\mathbf{CP} - \mathbf{RS} - \gamma   \mathbf{DQ}) \mathbf{P}_{m}^T \\
+&  \sum\limits_{n=1}^{N} [ 4 \mu (\sum\limits_{i \neq j}^M\mathbf{p}_{ni}^T\mathbf{C}_i^T\mathbf{C}_j\mathbf{p}_{nj} - \epsilon_1  )
(\sum\limits_{l=1, l \neq m}^M \mathbf{C}_l\mathbf{p}_{nl})\mathbf{p}_{nm}^T], \nonumber
\end{align}
where $\mathbf{P}_m = [\mathbf{p}_{1m},\cdots, \mathbf{p}_{Nm}]$.

\noindent \textbf{Update $\epsilon_1, \epsilon_2$.}
With other variables fixed,
it is easy to get the optimal solution,
\begin{align}
\epsilon_1^* = \frac 1 N \sum\nolimits_{n=1}^{N}\sum\nolimits_{i \neq j}^M \mathbf{p}_{ni}^T \mathbf{C}_{i}^T\mathbf{C}_{j}\mathbf{p}_{nj}, \\
\epsilon_2^* = \frac 1 N \sum\nolimits_{n=1}^{N}\sum\nolimits_{i \neq j}^M \mathbf{q}_{ni}^T \mathbf{D}_{i}^T\mathbf{D}_{j}\mathbf{q}_{nj}.
\end{align}

\noindent \textbf{Update $\mathbf{P}$ ($\mathbf{Q}$).}
The binary vectors $\{\mathbf{p}_n\}_{n=1}^N$
given other variables fixed
are independent with each other,
and hence the optimization problem
can be decomposed into $N$ sub-problems,
\begin{align}
\Psi_n =~& \|\mathbf{x}'_n - \mathbf{C}\mathbf{p}_n\|_2^2
+ \gamma \| \mathbf{C}\mathbf{p}_n - \mathbf{D}\mathbf{q}_n \|_2^2\\
+~& \mu (\sum\nolimits_{i\neq j}^M \mathbf{p}_{ni}^T \mathbf{C}_{i}^T\mathbf{C}_{j}\mathbf{p}_{nj} - \epsilon_1)^2.
\end{align}
This problem is a mixed-binary-integer problem
generally considered as NP-hard.
As a result, we approximately solve this problem by greedily updating the $M$ indicating vectors
$\{\mathbf{p}_{nm} \}_{m=1}^M$ in cycle:
fixing $\{\mathbf{p}_{nm'} \}_{m'=1,m'\neq m}^M$,
$\mathbf{p}_{nm}$ is updated by exhaustively checking all the elements in $\mathbf{C}_m$,
finding the element such that the objective function is minimized,
and setting the corresponding entry of $\mathbf{p}_{nm}$ to be 1 and all the others to be 0.
Similar optimization procedure is adopted to update $\mathbf{Q}$.


\subsection{Search process} \label{sec:searchprocess}
In cross-modal search, the given query can be either an image
or a text,
which require different querying processes.

\noindent \textbf{Image query.}
If the query is an image, $\mathbf{x}_q$, we first obtain the representation in the common space,
$\mathbf{x}_q' = \mathbf{Rs}^*$,
\begin{align}
\mathbf{s}^* = \text{arg}\min_{\mathbf{s}}~& \|\mathbf{x}_q - \mathbf{B} \mathbf{s}\|^2_2 + \rho |\mathbf{s}|_1.
\end{align}
The approximated distance between the
image query $\mathbf{x}_q$ and the database text
$\mathbf{y}_n$ (represented as $\mathbf{Dq}_n = \sum_{m=1}^M\mathbf{D}_m\mathbf{q}_{nm}$) is,
\begin{align}
\|\mathbf{x}_q' - \mathbf{Dq}_n\|_2^2 =
\sum\nolimits_{m=1}^M\|\mathbf{x}_q' - \mathbf{D}_m\mathbf{q}_{nm}\|_2^2\\
- (M-1)\|\mathbf{x}_q'\|_2^2
+ \sum\nolimits_{i\neq j}^M \mathbf{q}_{ni}^T\mathbf{D}_i^T\mathbf{D}_j\mathbf{q}_{nj}.
\end{align}
The last term
$\sum\nolimits_{i\neq j}^M \mathbf{q}_{ni}^T\mathbf{D}_i^T\mathbf{D}_j\mathbf{q}_{nj}$
is constant for all the texts
due to the introduced equality constraint
in Equation~\ref{eqn:constraint2}.
Hence given $\mathbf{x}_q'$, it is enough to compute
the first term $\sum\nolimits_{m=1}^M\|\mathbf{x}_q' - \mathbf{D}_m\mathbf{q}_{nm}\|_2^2$
to search for the nearest neighbors,
which furthermore can be efficiently computed and takes
$O(M)$ by
looking up a precomputed distance table
storing the distances: $\{\|\mathbf{x}_q' - \mathbf{d}_{mk}\|_2^2~|~m=1,\cdots,M;k=1,\cdots,K\}$.

\noindent \textbf{Text query.}
When the query comes as a text, $\mathbf{y}_q$,
the representation $\mathbf{y}_q'$ is obtained by solving,
\begin{align}
\mathbf{y}_q' = \text{arg}\min_{\mathbf{y}}~& \|\mathbf{y}_q - \mathbf{U} \mathbf{y}\|^2_2 .
\end{align}
Using $\mathbf{y}_q'$ to search in the image database
is similar to that in the image query search process.

\section{Discussions}

\noindent \textbf{Relation to compositional correlation quantization.}
The proposed approach is close to
compositional correlation quantization~\cite{LongWY15},
which is also a quantization-based method for cross-modal search.
In fact, our approach differs from it in two ways:
(1) we find a different mapping function to project the common space;
(2) we learn separate quantized centers for a pair using two dictionaries
instead of the unified quantized centers in
compositional correlation quantization~\cite{LongWY15}
imposed with a harder alignment using one dictionary.
Hence, during the quantization stage, our approach
can obtain potentially smaller quantization error,
as the quantized center
is more flexible,
and thus produce better search performance.
The empirical comparison illustrating the effect of dictionary
is shown in Figure~\ref{fig:samedictionary}.


\noindent \textbf{Relation to latent semantic sparse hashing.}
In our formulation, the common space is learnt in a similar
manner with latent semantic sparse hashing~\cite{ZhouDG14}.
After the common space mapping,
latent semantic sparse hashing applies
a simple sign function
directly on the common space,
which can result in large information loss
and hence weaken the search performance.
Our approach, however, adopts the quantization technique
that has more accurate distance approximation than hashing,
and produces better cross-modal search quality than latent semantic sparse hashing,
which is verified in our experiments
shown in Table~\ref{tab:map100all} and Figure~\ref{fig:prall}.

\begin{table*}[t] \scriptsize
\begin{center}
\caption{MAP$@50$ comparison of different algorithms on all the benchmark
datasets under various code lengths.
We also report the results of CMFH and CCQ (whose code implementations
are not publicly available) in their corresponding papers
and we distinguish those results by parenthesis $()$.
``---\hspace*{-3mm}---'' is used in the place where the result under that specific setting
is not reported in their papers.
Different setting refers to different datasets, or (and) different features, or (and) different bits, and so on.}
\label{tab:map100all}
\begin{tabular}{c|l|cccc|cccc|cccc}
\hline
\multirow{2}{*}{Task} & \multirow{2}{*}{Method} &
\multicolumn{4}{c|}{Wiki} & \multicolumn{4}{c|}{FLICKR$25K$} & \multicolumn{4}{c}{NUS-WIDE} \\
\cline{3-14}
& & 16 bits & 32 bits & 64 bits & 128 bits & 16 bits & 32 bits & 64 bits & 128 bits & 16 bits & 32 bits & 64 bits & 128 bits \\
\hline
\multirow{7}{0.2in}{Img to Txt}
& CMSSH~\cite{BronsteinBMP10} &  0.2110 &  0.2115
        & 0.1932 & 0.1909
        &  0.6468 & 0.6616
 & 0.6681 & 0.6624
 &  0.5243 &  0.5210
 &  0.5211 &  0.4813 \\
 & CVH~\cite{KumarU11} & 0.1947 &  0.1798
       & 0.1732 & 0.1912
       & 0.6450 & 0.6363
            &  0.6273 & 0.6204
            &  0.5352 & 0.5254
          &  0.5011 &  0.4705\\
  & MLBE~\cite{ZhenY12a} & \bf{0.3537} &  \bf{0.3947}
         & 0.2599 &  0.2247
         & 0.6085 &  0.5866
        & 0.5841 & 0.5883
        &  0.4472 &  0.4540
        &  0.4703 &  0.4026\\
   & QCH~\cite{WuYZWW15} & 0.1490 &  0.1726
         & 0.1621 &  0.1611
         &  0.5722 & 0.5780
        & 0.5618 & 0.5567
        &  0.5090 &  0.5270
          &  0.5208 &  0.5135 \\
    & LSSH~\cite{ZhouDG14} &  0.2396 &  0.2336
           &  0.2405 &  0.2373
           &  0.6328 &  0.6403
          &  0.6451 & 0.6511
          &  0.5368 &  0.5527
          &  0.5674 &  0.5723  \\
\cline{2-14}
     & CMFH~\cite{DingGZ14} & 0.2548 & 0.2591
            &  0.2594 &  \bf{0.2651}
            &  0.5886 & 0.6067
             & 0.6343 &  0.6550
             &  0.4740 & 0.4821
             &  0.5130 &  0.5068\\
                  & (CMFH~\cite{DingGZ14}) & (0.2538) & (0.2582)
                         &  (\bf{0.2619}) &  (0.2648)
                         &  ---\hspace*{-2mm}--- &  ---\hspace*{-2mm}---
                          & ---\hspace*{-2mm}--- &  ---\hspace*{-2mm}---
                          &  ---\hspace*{-2mm}---& ---\hspace*{-2mm}---
                          &  ---\hspace*{-2mm}--- & ---\hspace*{-2mm}---\\
   \cline{2-14}
   & (CCQ~\cite{LongWY15}) & (0.2513) & (0.2529)
                            &  (0.2587) &  ---\hspace*{-2mm}---
                            &  ---\hspace*{-2mm}--- &  ---\hspace*{-2mm}---
                             & ---\hspace*{-2mm}--- &   ---\hspace*{-2mm}---
                             &  ---\hspace*{-2mm}--- & ---\hspace*{-2mm}---
                             &  ---\hspace*{-2mm}--- &  ---\hspace*{-2mm}--- \\
     \cline{2-14}
      & CMCQ &  0.2478 &  0.2513
            & 0.2567 &  0.2614
             &  \textbf{0.6705 }& \textbf{ 0.6716 }
             & \textbf{0.6782} & \textbf{0.6821}
             &  \textbf{0.5637 }&  \textbf{0.5902 }
             & \textbf{ 0.5990} &  \textbf{0.6096}\\
\hline
\hline
\multirow{7}{0.2in}{Txt to Img}
& CMSSH~\cite{BronsteinBMP10} &  0.2446 &  0.2505
        &  0.2387 &  0.2352
        &  0.6123 &  0.6400
       &  0.6382 &  0.6242
        &  0.4177 &  0.4259
       &  0.4187 & 0.4203\\
 & CVH~\cite{KumarU11} & 0.3186 & 0.2354
       & 0.2046 &  0.2085
       &  0.6595 & 0.6507
              &  0.6463 &  0.6580
              &  0.5601 &  0.5439
             &  0.5160 & 0.4821\\
  & MLBE~\cite{ZhenY12a} &  0.3336 &  0.3993
         &  0.4897 &  0.2997
         &  0.5937 &  0.6182
          &  0.6550 &  0.6392
          &  0.4352 & 0.4888
          &  0.5020 &  0.4425\\
   & QCH~\cite{WuYZWW15} & 0.1924 & 0.1561
         &  0.1800 &  0.1917
         &  0.5752 &  0.6002
          &  0.5757 &  0.5723
          &  0.5099 &  0.5172
         &  0.5092 & 0.5089\\
    & LSSH~\cite{ZhouDG14} &  0.5776 &  0.5886
           & 0.5998 & 0.6103
           &  0.6504 &  0.6726
           &  0.6965 &  0.7010
           &  0.6357 &  0.6638
            &  0.6820 &  0.6926\\
                  \cline{2-14}
     & CMFH~\cite{DingGZ14} &  0.6153 & 0.6363
            &  0.6411 & 0.6504
            &  0.5873 &  0.6019
           & 0.6477 & 0.6623
           &  0.5109 &  0.5643
           &  0.5896 &  0.5943\\
                         & (CMFH~\cite{DingGZ14}) & (0.6116) & (0.6298)
                                  &  (0.6398) &  (0.6477)
                                    &  ---\hspace*{-2mm}--- &  ---\hspace*{-2mm}---
                                     & ---\hspace*{-2mm}--- &   ---\hspace*{-2mm}---
                                     & ---\hspace*{-2mm}--- & ---\hspace*{-2mm}---
                                     &  ---\hspace*{-2mm}---& ---\hspace*{-2mm}---\\
              \cline{2-14}
              & (CCQ~\cite{LongWY15}) & (0.6351) & (0.6394)
                                       &  (0.6405) &  ---\hspace*{-2mm}---
                                       &  ---\hspace*{-2mm}--- & ---\hspace*{-2mm}---
                                        & ---\hspace*{-2mm}--- &   ---\hspace*{-2mm}---
                                        &  ---\hspace*{-2mm}--- & ---\hspace*{-2mm}---
                                        &  ---\hspace*{-2mm}--- &  ---\hspace*{-2mm}--- \\
               \cline{2-14}
      & CMCQ &  \bf{0.6397} &  \bf{0.6474}
            & \bf{0.6546} &  \bf{0.6593}
            & \textbf{0.7248} & \textbf{ 0.7335}
            & \textbf{0.7394} &  \textbf{0.7550}
            &  \textbf{0.6898} &  \textbf{0.7086}
            &  \textbf{0.7194} &  \textbf{0.7254}\\
\hline
\end{tabular}
\end{center}
\vspace{-0.9cm}
\end{table*}

\section{Experiments}

\subsection{Setup}

\noindent \textbf{Datasets.}
We evaluate our method on three benchmark datasets.
The first dataset, \textbf{Wiki}\footnote{http://www.svcl.ucsd.edu/projects/crossmodal/}
consists of 2,866 images and
2,866 texts describing the images in short paragraph (at least 70 words),
with images represented as 128-dimensional SIFT features
and texts expressed as 10-dimensional topics vectors.
This dataset is divided into 2,173 image-text pairs and 693 quries, and
each pair is labeled with one of the 10 semantic classes.
The second dataset,
\textbf{FLICKR25K}\footnote{http://www.cs.toronto.edu/~nitish/multimodal/index.html},
is composed of 25,000
images along with the user assigned tags. The average number of tags for an image is 5.15~\cite{SrivastavaS14}.
Each image-text pair is assigned with multiple labels from a total of 38 classes.
As in~\cite{SrivastavaS14},
the images are represented by 3857-dimensional features
and the texts are 2000-dimensional vectors indicating the occurrence of the tags. We randomly sampled $10\%$ of the pairs as the test set and use the remaining as the training set.
The third dataset is \textbf{NUS-WIDE}\footnote{http://lms.comp.nus.edu.sg/research/NUS-WIDE.htm}~\cite{nus-wide-civr09}
containing 269,648 images with associated tags (6 in average), each pair is annotated with multiple labels among 81 concepts.
As done in previous work~\cite{DingGZ14,ZhenY12b,ZhouDG14},
we select 10 most popular concepts resulting in 186,577 data pairs.
The images are represented by 500-dimensional bag-of-words features
based on SIFT descriptors,
and the texts are 1000-dimensional vectors of the most frequent tags.
Following~\cite{ZhouDG14},
We use 4000 ($\approx2\%$) randomly sampled pairs as the query set and the rest as the training set.

\noindent \textbf{Evaluation.}
In our experiments, we report the results of two search tasks for the cross-modal search, i.e., the image (as the query) to text (as the database) task and
the text to image task.
The search quality is evaluated with two measures:
MAP$@T$ and precision$@T$.
MAP$@T$ is defined as the mean of the average precisions of all the queries,
and the average precision of a query is computed as,
$AP(\mathbf{q}) =
\frac {\sum_{t=1}^T P_{\mathbf{q}}(t)\delta(t)}
{\sum_{t=1}^T\delta(t)}$,
where $T$ is the number of retrieved items,
$P_{\mathbf{q}}(t)$ is the precision at position $t$ for query $\mathbf{q}$,
and $\delta(t) = 1$ if the retrieved $t$th item has the same label with query $\mathbf{q}$ or shares at least one label,
otherwise $\delta(t)=0$.
Following~\cite{ZhouDG14,DingGZ14,LongWY15},
we report MAP$@T$ with $T=50$ and $T=100$.
We also plot the precision$@T$ curve
which is obtained by
computing the precisions at different recall levels
through varying the number of retrieved items.

\noindent \textbf{Compared methods.}
We compare our approach,
Cross-Modal Collaborative Quantization (CMCQ), with
three baseline methods that only use the intra-document relation:
Latent Semantic Sparse Hashing (LSSH)~\cite{ZhouDG14},
Collective Matrix Factorization Hashing (CMFH)~\cite{DingGZ14},
and Compositional Correlation Quantization (CCQ)~\cite{LongWY15}.
The code of LSSH is generously provided by the authors
and we implemented the CMFH carefully by ourselves.
The performance of CCQ (without public code) is presented partially using the results in its paper.
In addition,
we report the state-of-the-art algorithms
whose codes are publicly available:
(1) Cross-Modal Similarity Sensitive Hashing (CMSSH)~\cite{BronsteinBMP10},
(2) Cross-View Hashing (CVH)~\cite{KumarU11},
(3) Multimodal Latent Binary Embedding (MLBE)~\cite{ZhenY12a},
(4) Quantized Correlation Hashing (QCH)~\cite{WuYZWW15}.
The parameters in above methods
are set according to the corresponding papers.

\noindent \textbf{Implementation details.}
The data for both modalities are mean-centered
and then
normalized to have unit Euclidean length.
We use principle component analysis to project the image
into a lower dimensional (set to 64) space,
and the number of bases in sparse coding is set to 512 ($\mathbf{B}\in \mathbb{R}^{64 \times 512}$).
The latent dimension of matrix
factorization for text data is
set equal to the number of code bits, e.g., 16, 32 etc.
The mapping parameters
(denoted as $\boldsymbol{\theta}_m$)
are initialized by solving a
relatively easy problem
$\min \mathcal{M}(\boldsymbol{\theta}_m)$
(similar algorithm with that presented in solving $\min \mathcal{F}(\boldsymbol{\theta}_m|\boldsymbol{\theta}_q)$).
Then the quantization parameters (denoted as $\boldsymbol{\theta}_q$) are initialized
by conducting composite quantization~\cite{ZhangDW14}
in the common space.

There are five parameters balancing different trade-offs
in our algorithm:
the sparsity degree $\rho$,
the scale-balance parameter $\eta$,
the alignment degree in the common space $\lambda$,
the correlation degree of the quantization $\gamma$,
and the penalty parameter $\mu$.
We simply set $\mu = 0.1$ in our experiments as it
has already shown satisfactory results.
The other four parameters are selected through validation
(by varying one parameter in $\{0.1,0.3,0.5,0.7\}$
while keeping others fixed)
so that
the MAP value, when using the validation set (a subset of the training data) as the queries to search in the remaining training data, is the best.
The sensitive analysis of these parameters is presented in Section~\ref{sec:parameteranalysis}.

\subsection{Results}

\begin{figure*}[t]
\centering
(a)\includegraphics[width=.22\linewidth, clip]{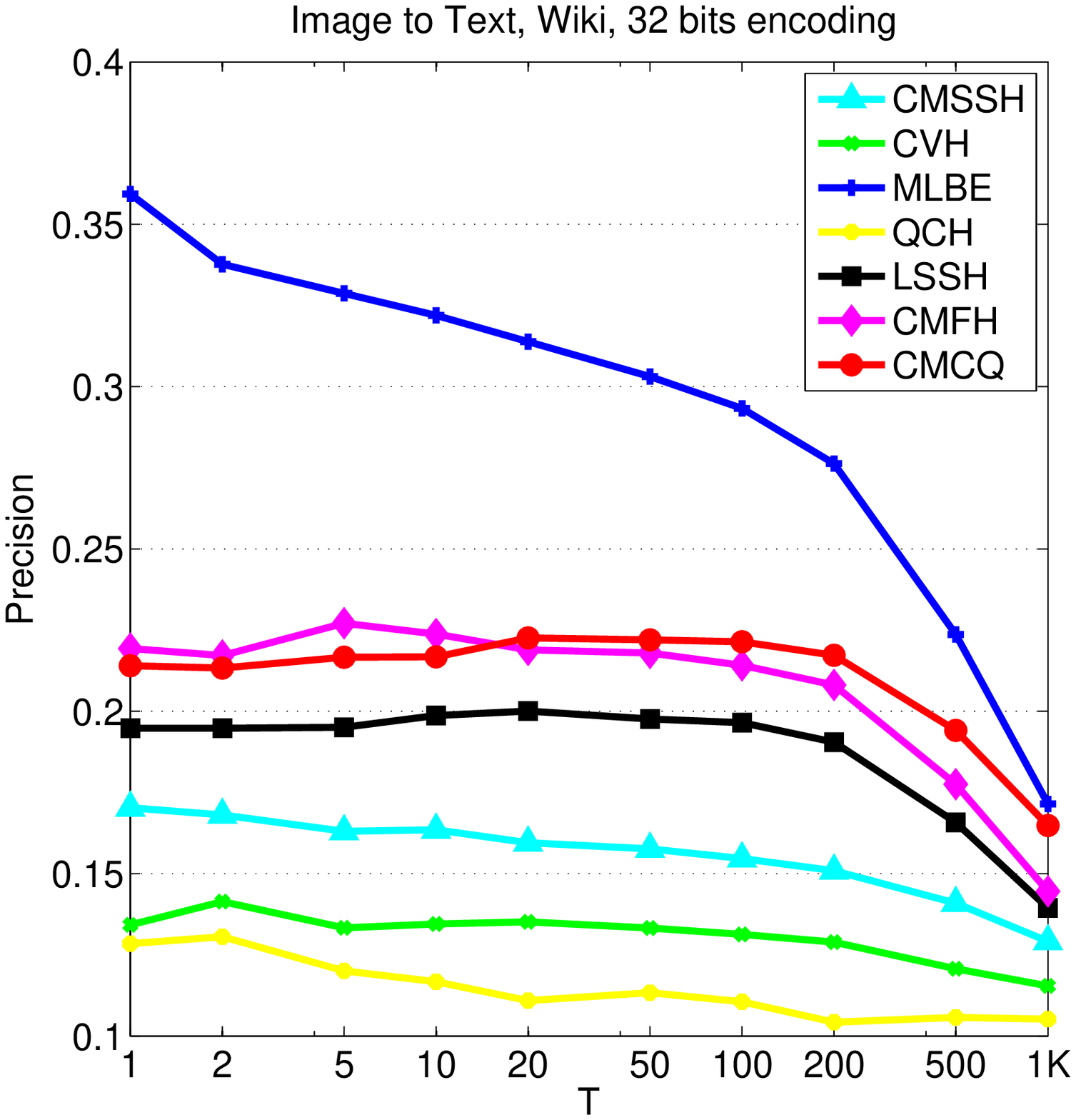}~
\includegraphics[width=.22\linewidth, clip]{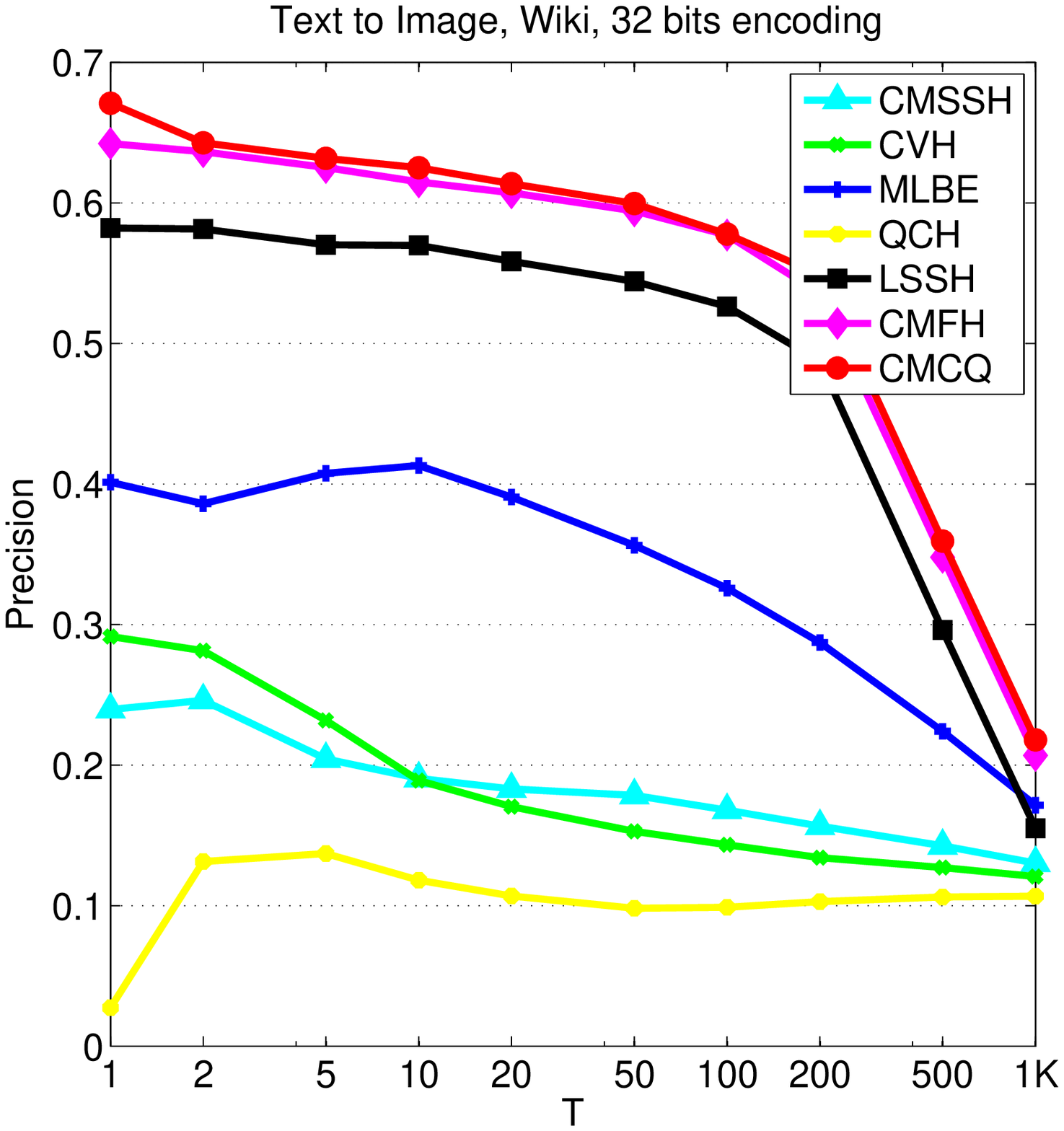}~
\includegraphics[width=.22\linewidth, clip]{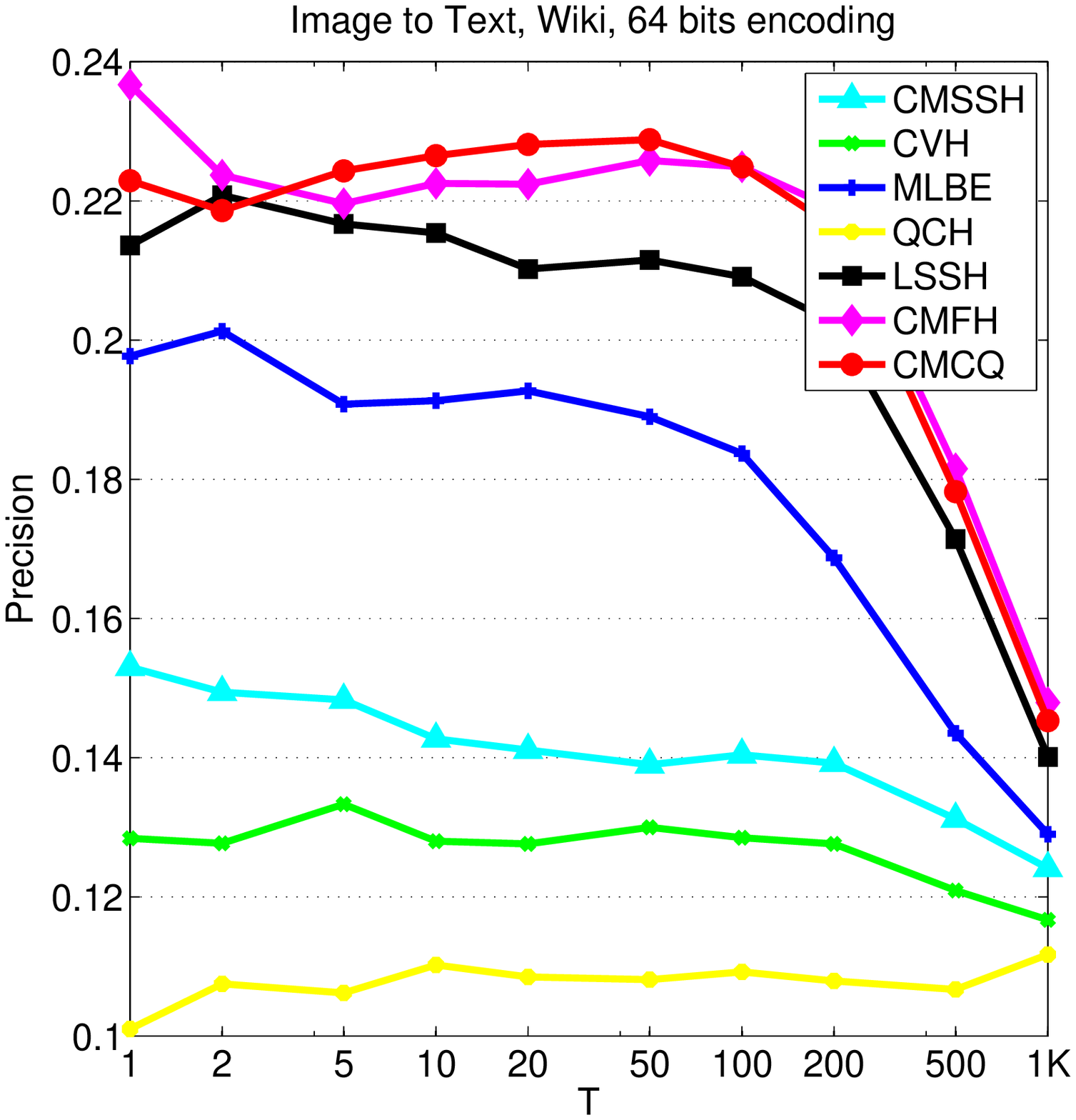}~
\includegraphics[width=.22\linewidth, clip]{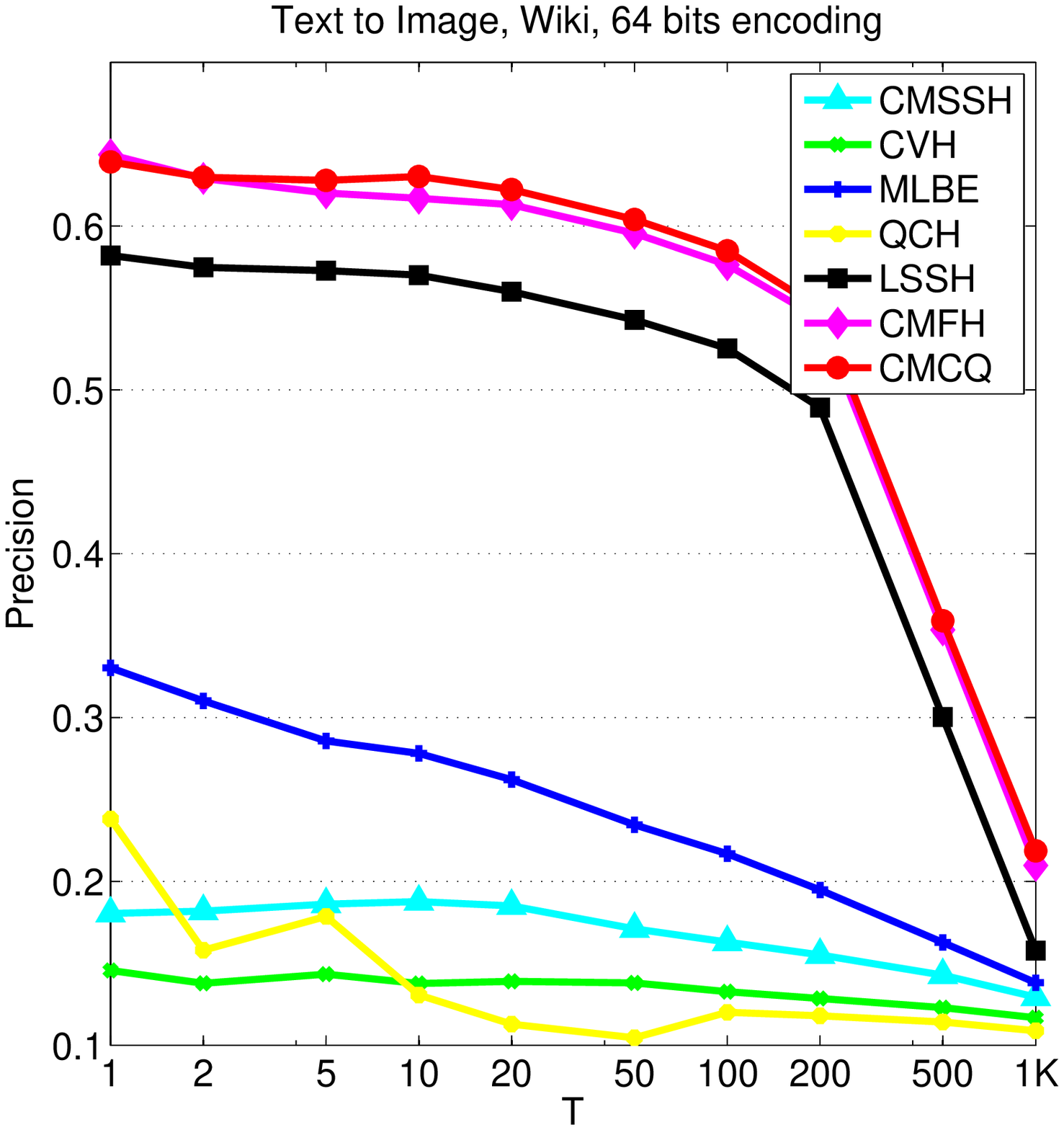}

(b)\includegraphics[width=.22\linewidth, clip]{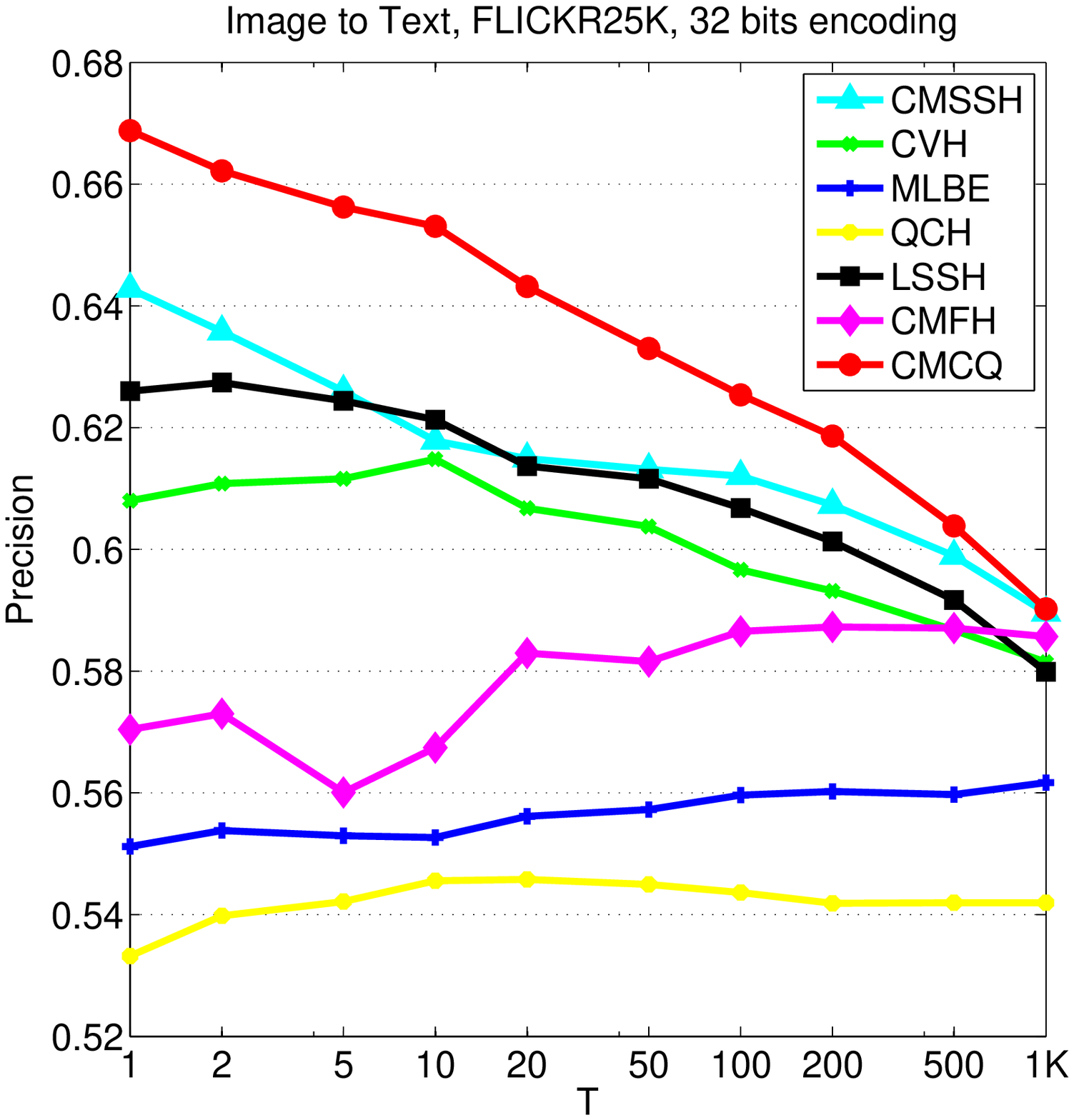}~
\includegraphics[width=.22\linewidth, clip]{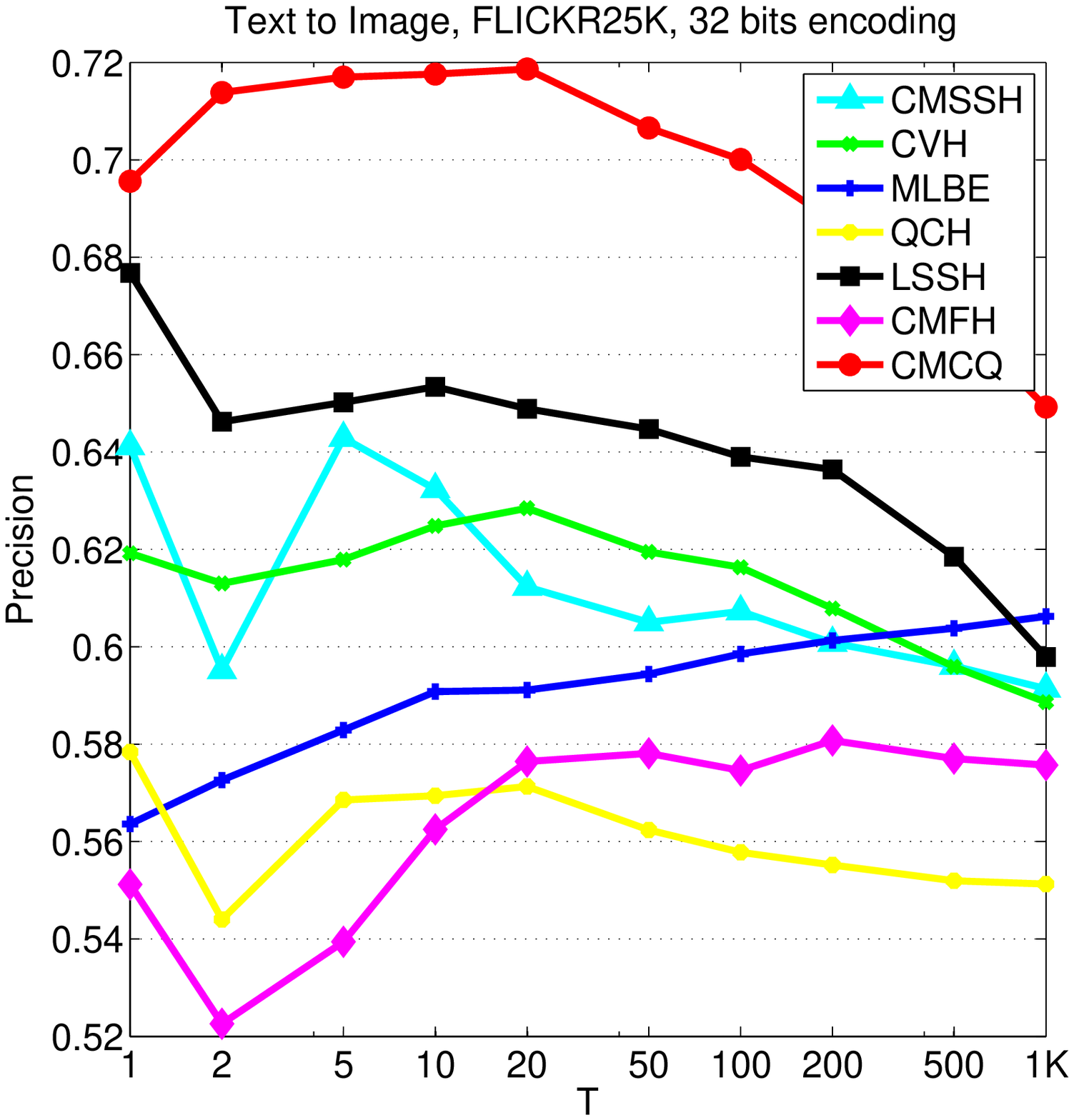}~
\includegraphics[width=.22\linewidth, clip]{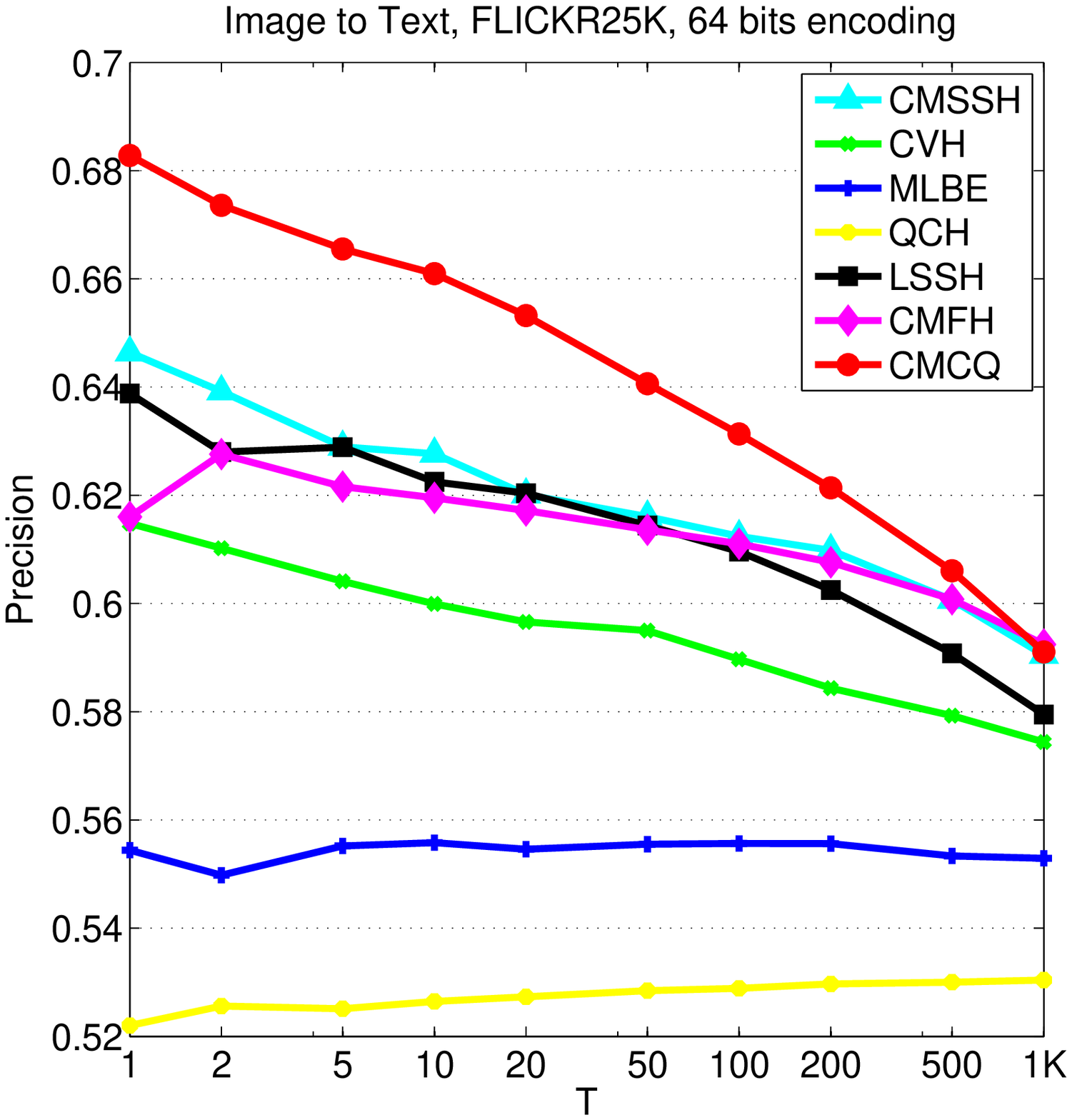}~
\includegraphics[width=.22\linewidth, clip]{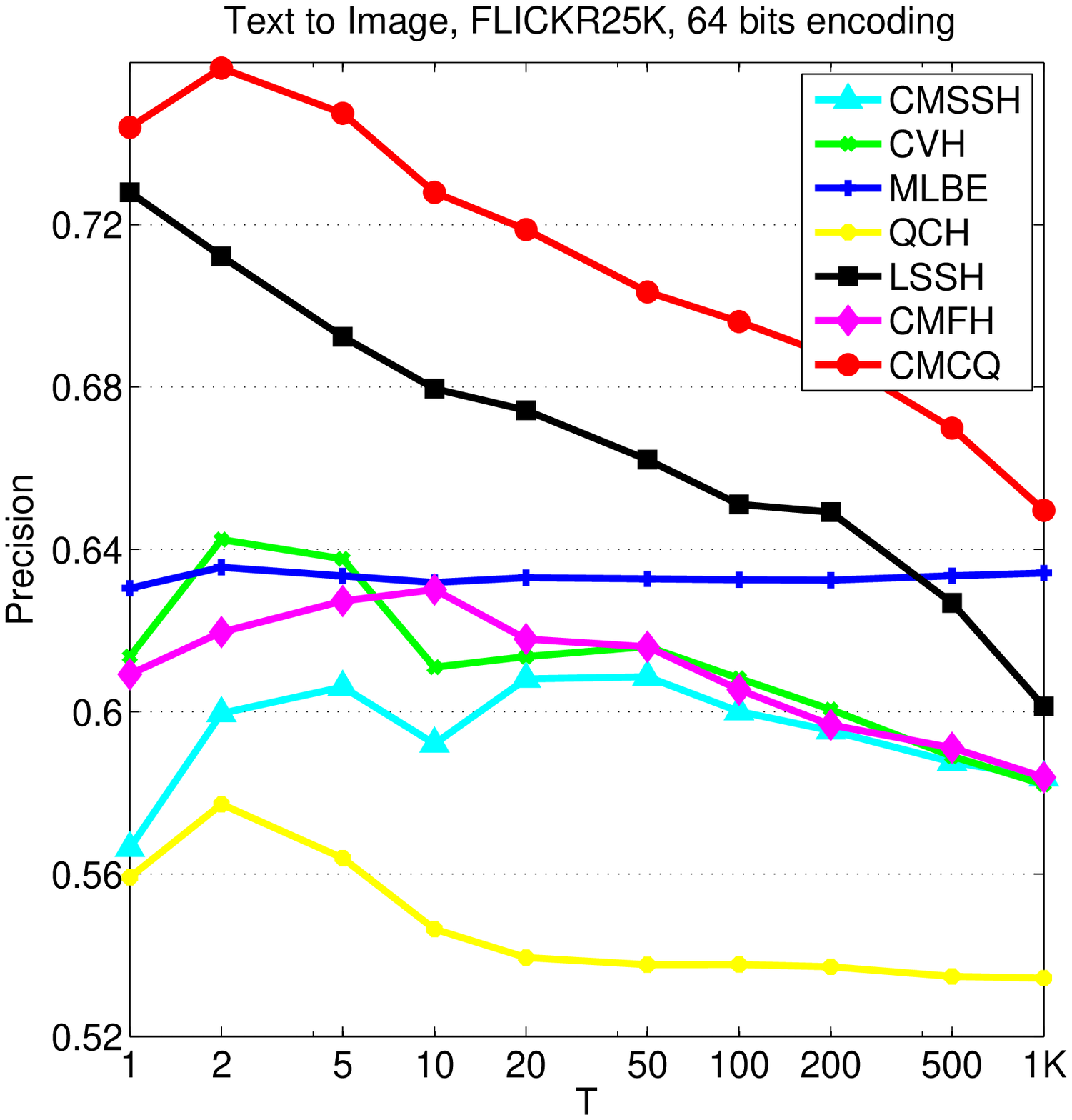}

(c)\includegraphics[width=.22\linewidth, clip]{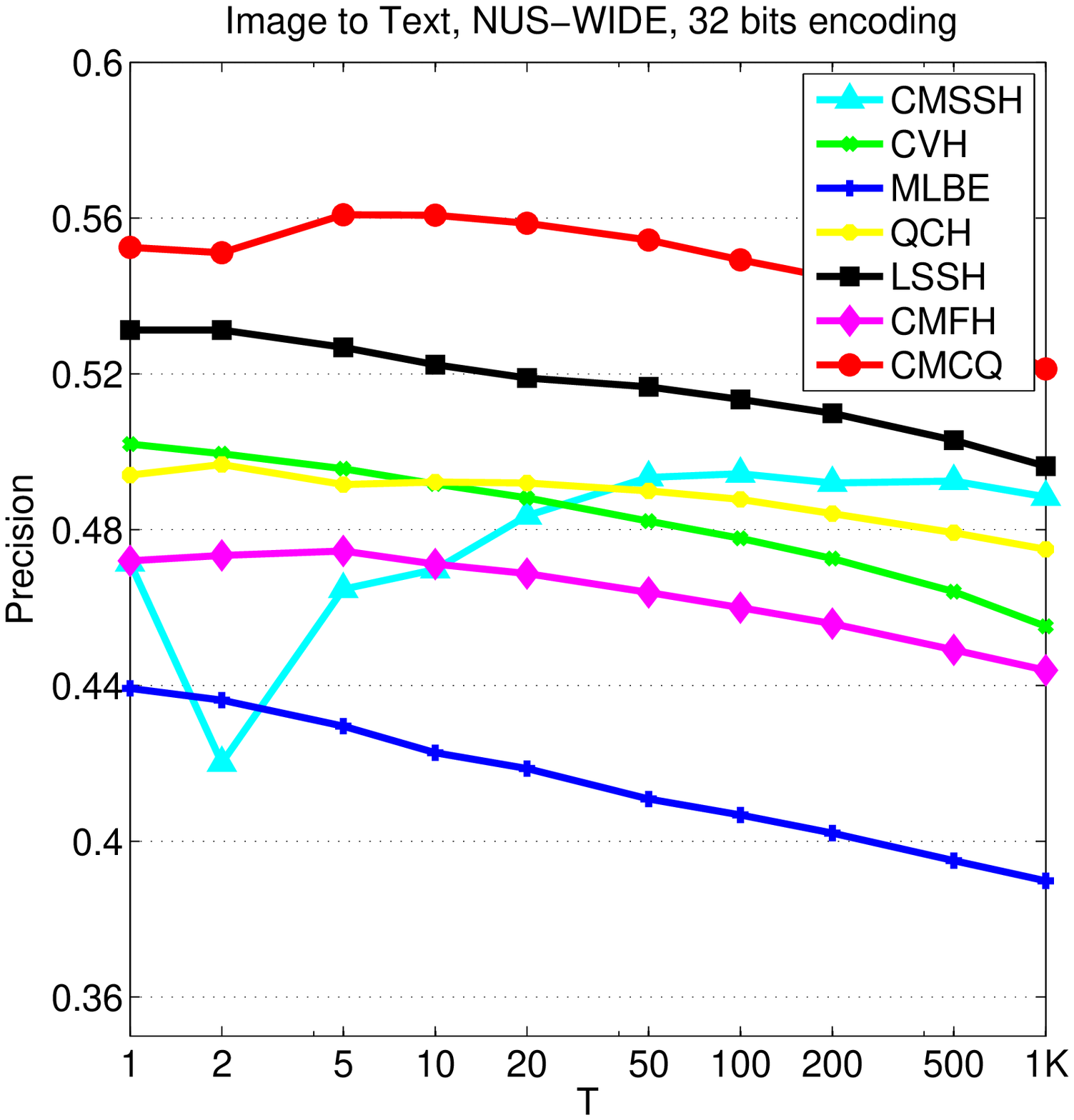}~
\includegraphics[width=.22\linewidth, clip]{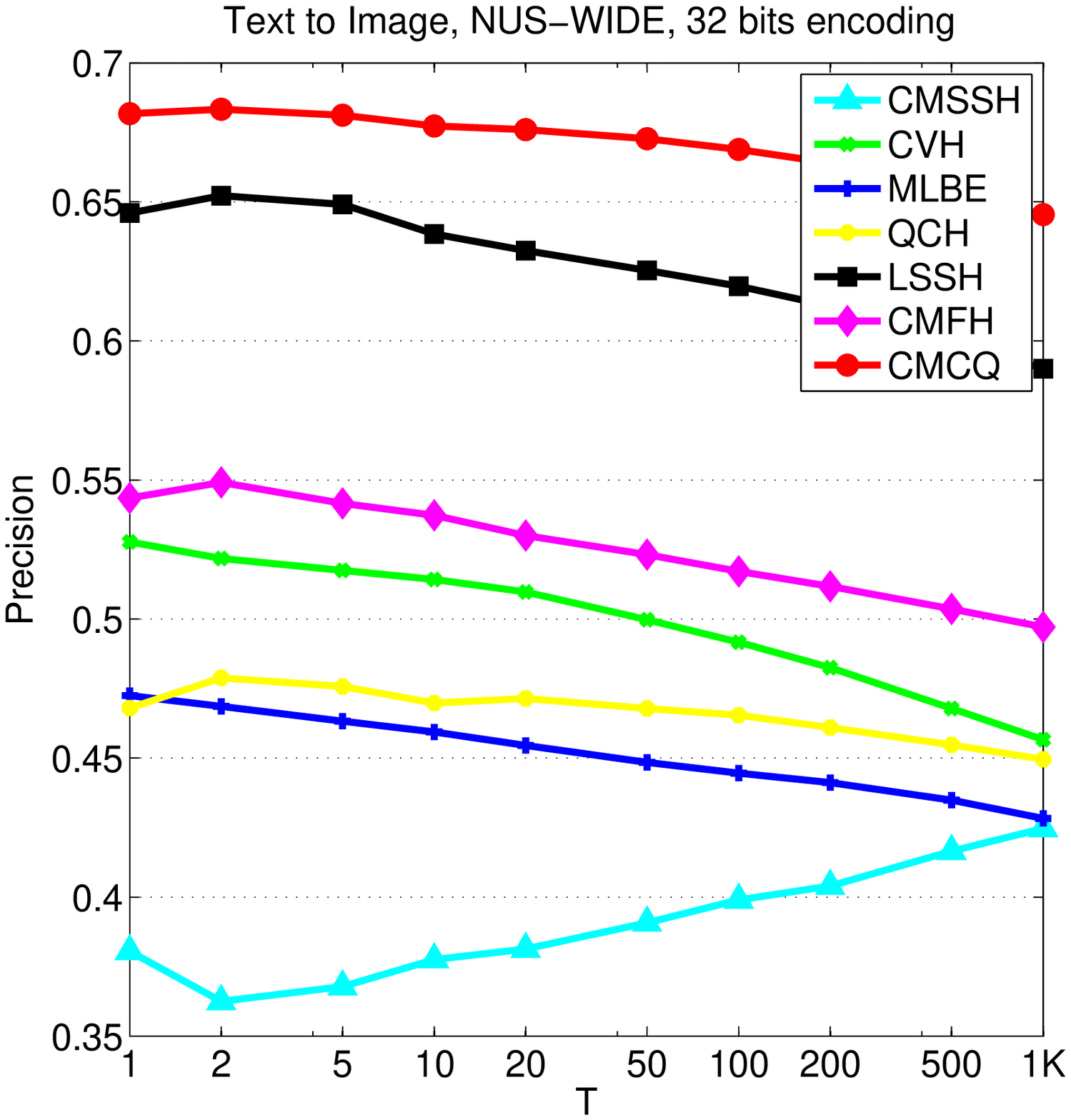}~
\includegraphics[width=.22\linewidth, clip]{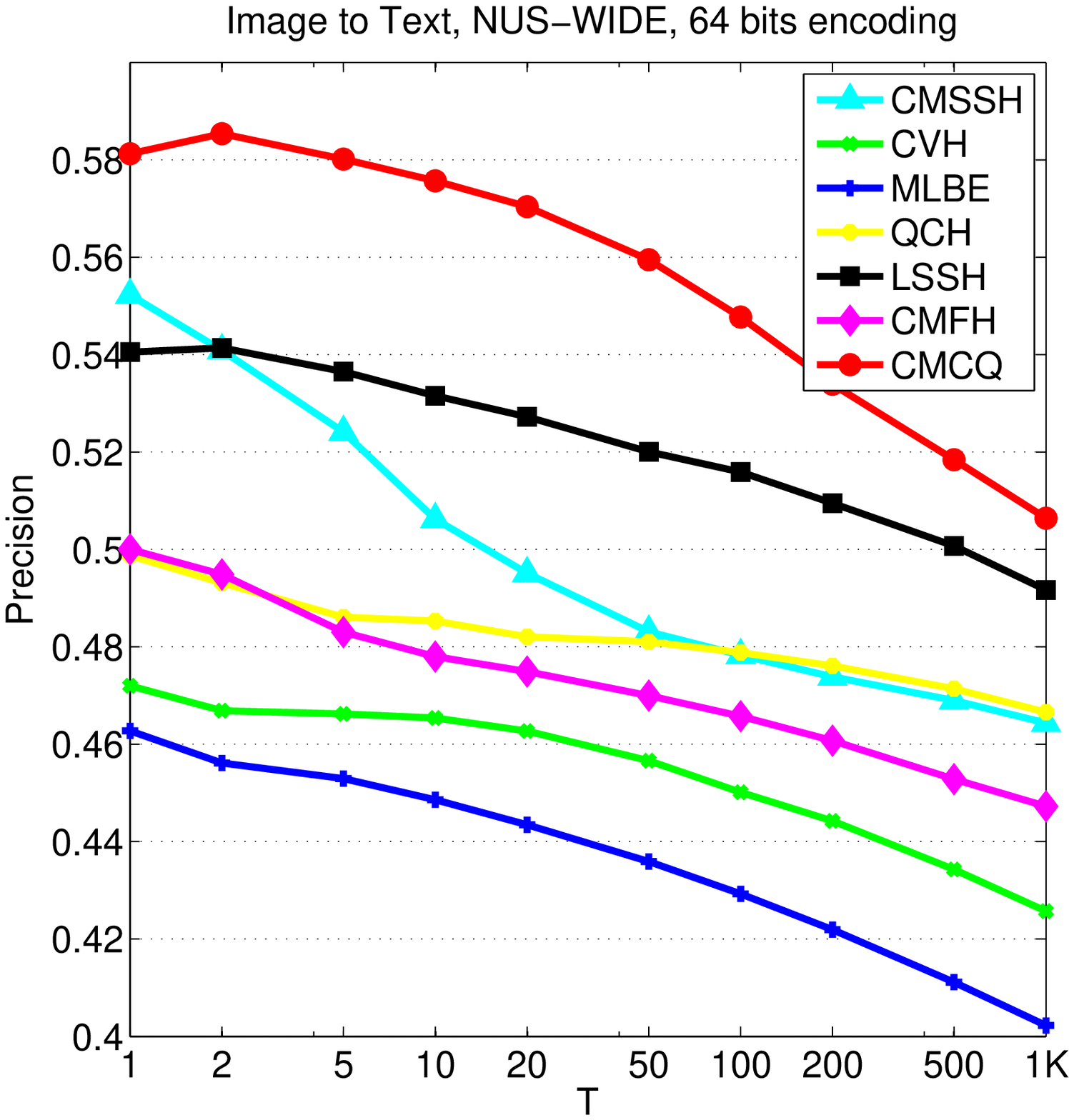}~
\includegraphics[width=.22\linewidth, clip]{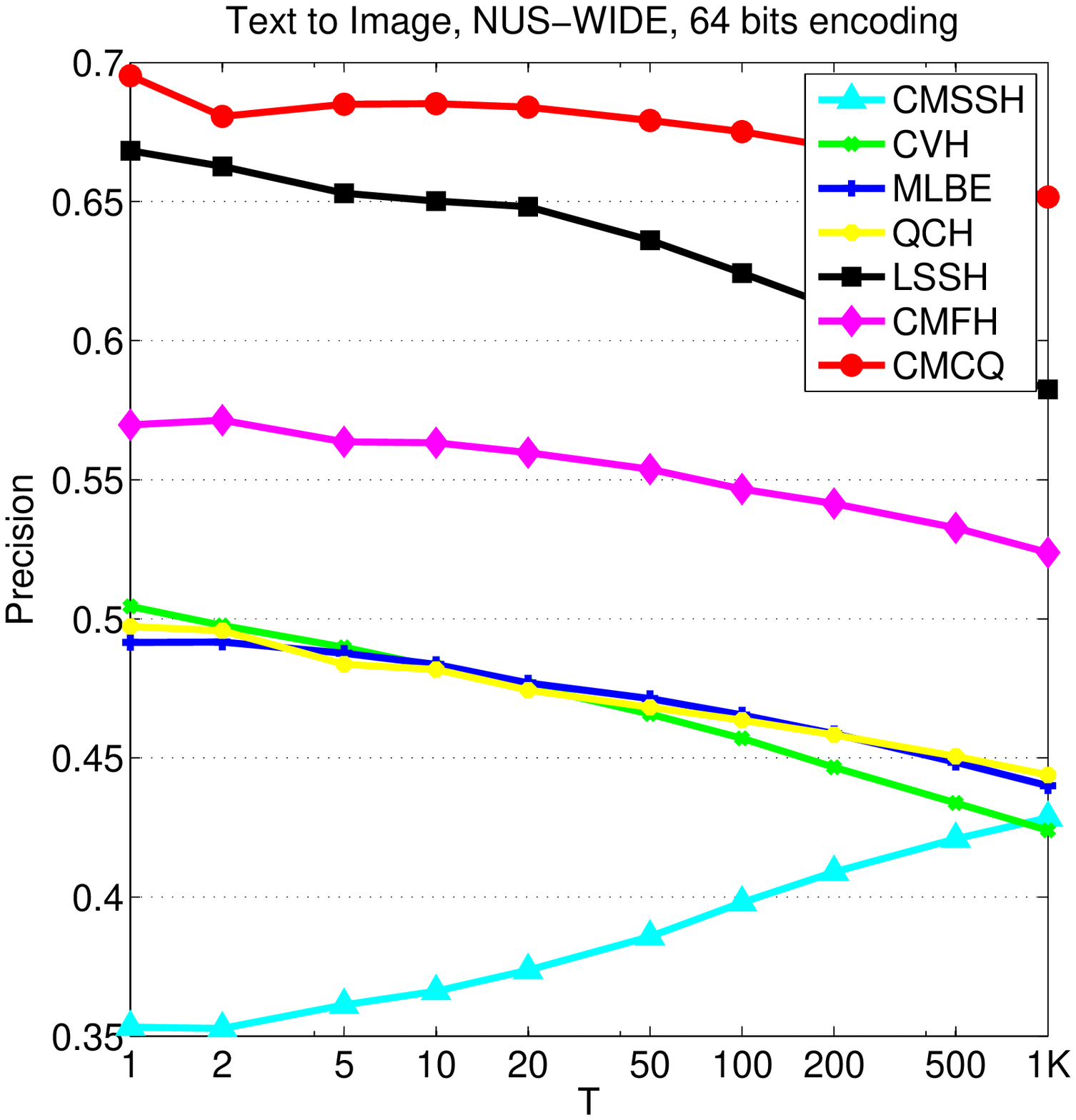}

\caption{Precision$@T$ ($T$ is the number of retrieved items) curve of different algorithms
 on the (a) Wiki, (b) FLICKR$25K$, and (c) NUS-WIDE dataset encoded with 32 bits and 64 bits over two search tasks:
 image to text
 and text to image.}
\label{fig:prall}
\vspace{-0.3cm}
\end{figure*}

\noindent \textbf{Results on Wiki.}
The comparison in terms of MAP@$100$ and the precision$@T$ curve is reported in Table~\ref{tab:map100all} and the first row of Figure~\ref{fig:prall}.
We can find that our approach, CMCQ, achieves
better performance than other methods over the text to image task.
While over the image to query task,
we can see from Table~\ref{tab:map100all} that
the best performance is achieved by
MLBE with 16 bits and 32 bits,
and CMFH with 64 bits and 128 bits.
However, the performance of MLBE decreases as the code length
gets longer.
Our approach, on the other hand, is able
to utilize the additional bits to
enhance the search quality.
In comparison with CMFH,
we can see that
our approach gets the similar results.

\noindent \textbf{Results on FLICKR25K.}
The performance on the FLICKR25K dataset is shown in
Table~\ref{tab:map100all} and the
second row of Figure~\ref{fig:prall}.
It can be seen that
the gain obtained by our approach
is significant over both cross-modal search tasks.
Moreover, we can observe from Table~\ref{tab:map100all}
that the results of our approach with the smallest
code bits perform much better than other methods
with the largest code bits.
For example, over the text to image task,
the MAP$@50$ of our approach, CMCQ with $16$ bits, is $0.7248$,
about $2\%$ larger than $0.7010$, the best MAP$@50$
obtained by
other baseline methods with $128$ bits.
This indicates that when dealing with high-dimensional dataset, such as FLICKR$25K$ with $3857$-dimensional image features and $2000$-dimensional text features,
our method
keeps much more information than other hashing-based
cross-modal techniques,
and hence produces better search quality.

\noindent \textbf{Results on NUS-WIDE.}
Table~\ref{tab:map100all} and the third row of Figure~\ref{fig:prall}
show the performance of all the methods on the largest dataset of the three datasets, NUS-WIDE.
One can observe that the proposed approach again gets
the best performance.
In addition,
it can be seen from the figure that
in most cases,
the performance of our approach barely drops with increasing
value of $T$.
For instance, the precision$@1$ of our approach
over the text to image task with
$32$ bits is $68.17\%$, and the precision$@1K$
is $64.55\%$,
which suggests that our method consistently keeps
a large portion of the relevant items retrieved
as the number of retrieved items increases.

\subsection{Empirical analysis} \label{sec:parameteranalysis}

\noindent \textbf{Comparison with semantics-preserving hashing.}
Another challenging competitor for our approach is the recent
semantics-preserving hashing (SePH)~\cite{LinDH015}.
The comparison is shown in Table~\ref{tab:SPH} (A).
The reason of SePH outperforming ours is 
that
SePH exploits the document-label information,
which our method doesn't use for two reasons:
(1) the image-text correspondence information
comes naturally and easily,
while the label information is expensive to get;
(2) exploiting label information may tend to overfit the data
and not generalize well to newly-coming classes.
To show it,
we conducted an experiment:
split the NUS-WIDE training set into two parts:
one with five concepts for training,
and the other with other five concepts
for the search database
whose codes are extracted using the model trained on the first part. 
Our results as shown in Table~\ref{tab:SPH} (B)
are better than SePH,
indicating that our method can well generalize to newly-coming classes.

\begin{table}[t] \scriptsize
\begin{center}
\caption{Comparison with SePH (in terms of MAP$@50$).}
\label{tab:SPH}
\begin{tabular}{c||c|c|ccc}
\hline
\multicolumn{6}{c}{\textbf{(A)} Comparison between ours and SePH$_{km}$ (the best version of SePH)}\\
\hline
 Dataset & Task & Method & $ 32 $ & $ 64 $ & $ 128 $ \\
  \hline
  \multirow{4}{*}{NUS-WIDE} & \multirow{2}{*}{Img to Txt} & SePH$_{km}$
   &  $ 0.586 $ & $ 0.601 $ & $ 0.607 $ \\
   & & CMCQ & $ 0.590 $ & $ 0.599 $　& $ 0.610 $ \\
   \cline{2-6}
   & \multirow{2}{*}{Txt to Img} & SePH$_{km}$
    &  $ 0.726 $ & $ 0.746 $ & $ 0.746 $ \\
   & & CMCQ &  $ 0.709 $ & $ 0.719 $　& $ 0.725 $ \\
\hline
\multicolumn{6}{c}{\textbf{(B)} Generalization to \emph{``newly-coming classes"}: ours outperforms SePH$_{km}$}\\
\hline
 \multirow{4}{*}{NUS-WIDE} & \multirow{2}{*}{Img to Txt} & SePH$_{km}$
 &$  0.442 $ & $ 0.436 $ & $ 0.435 $ \\
  && CMCQ & $ 0.495 $ & $ 0.501 $　& $ 0.535 $ \\
\cline{2-6}
 &\multirow{2}{*}{Txt to Img} & SePH$_{km}$
  &  $ 0.448 $ & $ 0.459 $ & $ 0.455 $ \\
 && CMCQ &  $ 0.551 $ & $ 0.568 $　& $ 0.580 $ \\
   \hline
\end{tabular}
\end{center}
\vspace{-0.9cm}
\end{table}

\begin{figure}[t]
\centering
~~\includegraphics[width = 0.48\linewidth, clip]{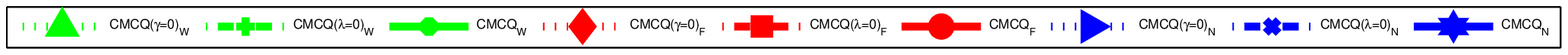}~~\\
~~\includegraphics[width = 0.48\linewidth, clip]{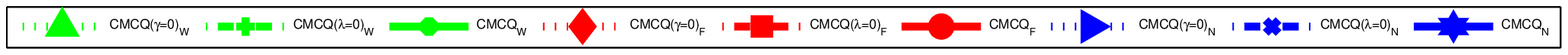}~~\\
~~\includegraphics[width = 0.48\linewidth, clip]{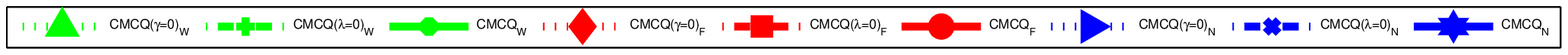}~~\\
\includegraphics[width=.46\linewidth, clip]{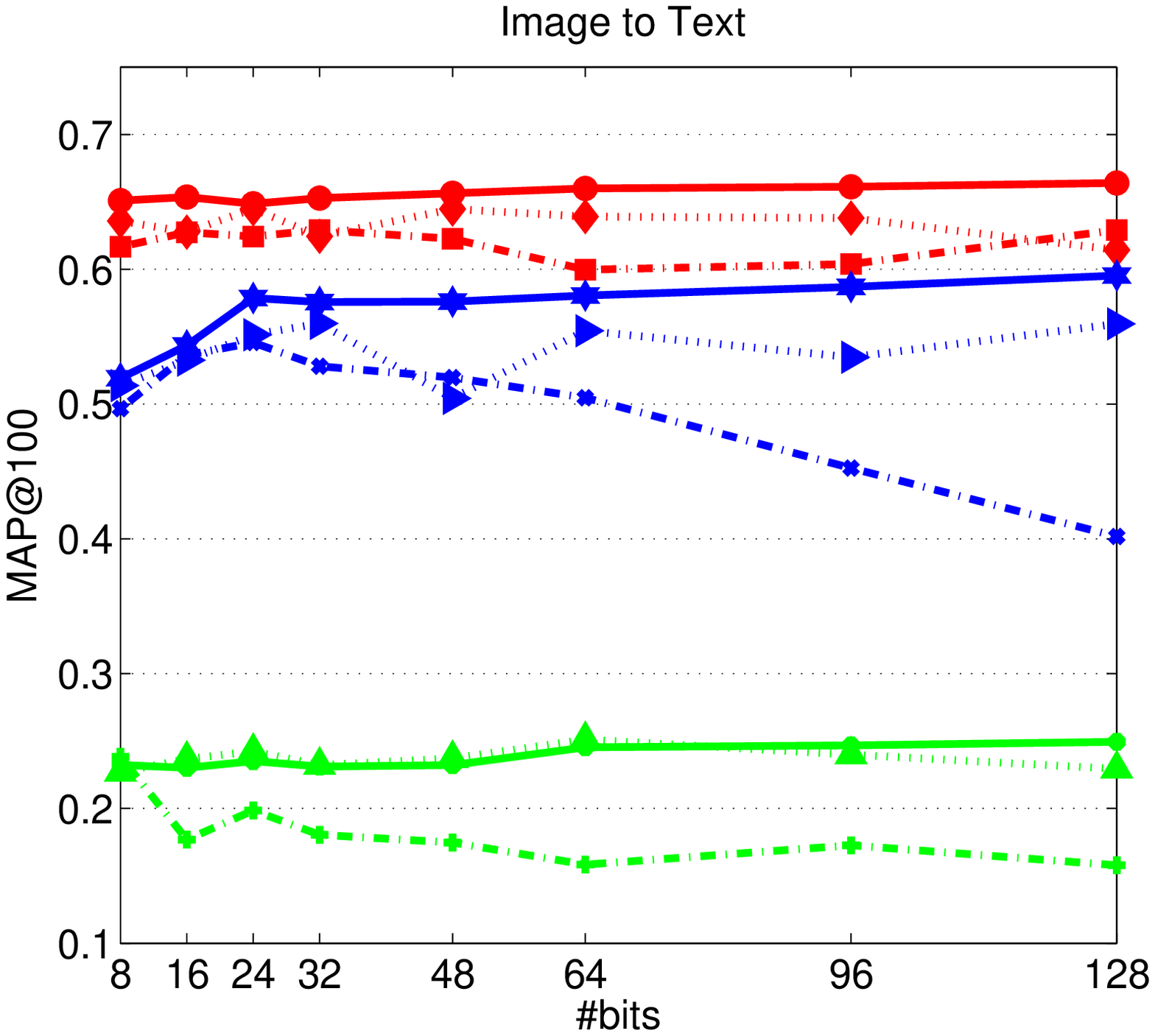}~
\includegraphics[width=.46\linewidth, clip]{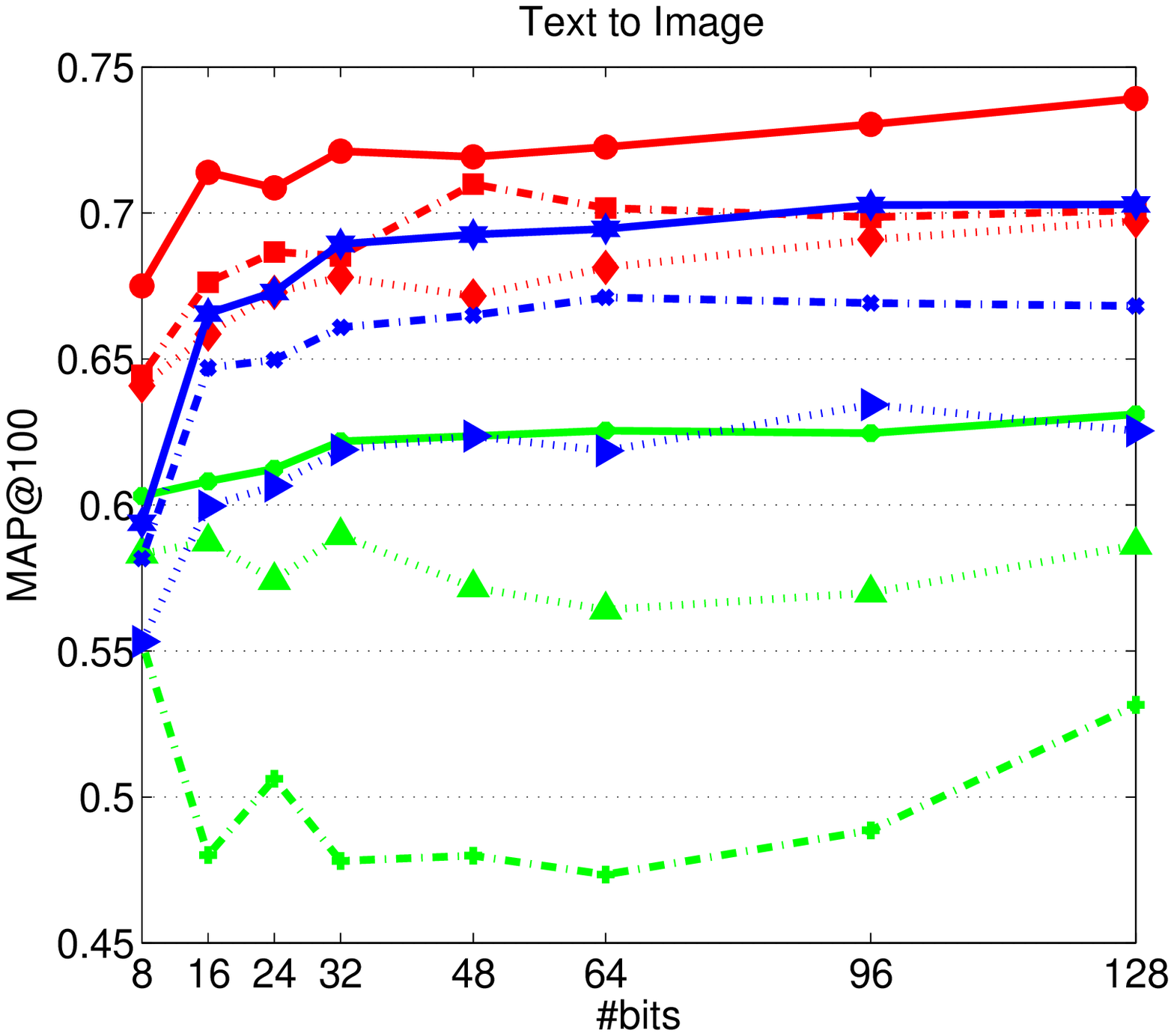}
\caption{Illustrating the effect of the intra-document relation. The MAP is compared
among CMCQ, CMCQ ($\gamma=0$) (without correlation in
the quantized space), and CMCQ ($\lambda=0$)
(without correlation in the common space)
on the three datasets denoted as
W (Wiki), F (FLICKR$25K$), and N (NUS-WIDE) in the legend.}
\label{fig:gamma0lambda0}
\vspace{-0.3cm}
\end{figure}

\noindent \textbf{The effect of intra-document correlation.}
The intra-document correlation
is imposed
in our formulation
over two spaces
(the quantized space and the common space)
by two
regularization terms controlled respectively by parameter $\gamma$ and $\lambda$.
In fact, it is possible to just add one such term and set the other to be $0$.
Specifically,
if $\gamma=0$,
our approach
will degenerate to conducting composite quantization~\cite{ZhangDW14}
separately on each modality,
and if $\lambda=0$,
the proposed approach will lack the explicit connection in the common space.
In either case, the bridge
that links the pair of image and text would be undermined,
resulting in reduced cross-modal search quality.
The experimental results
shown in
Figure~\ref{fig:gamma0lambda0},
validate this point:
the performance of our approach when considering
both of the intra-document correlation terms
is much better.

\begin{figure}[t]
\centering
~~\includegraphics[width = .98\linewidth, clip]{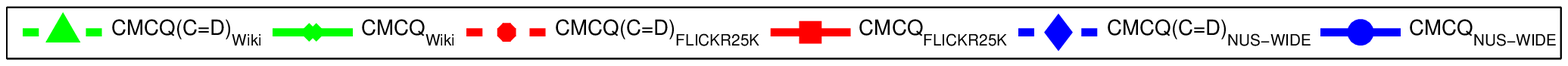}~~\\
\includegraphics[width=.46\linewidth, clip]{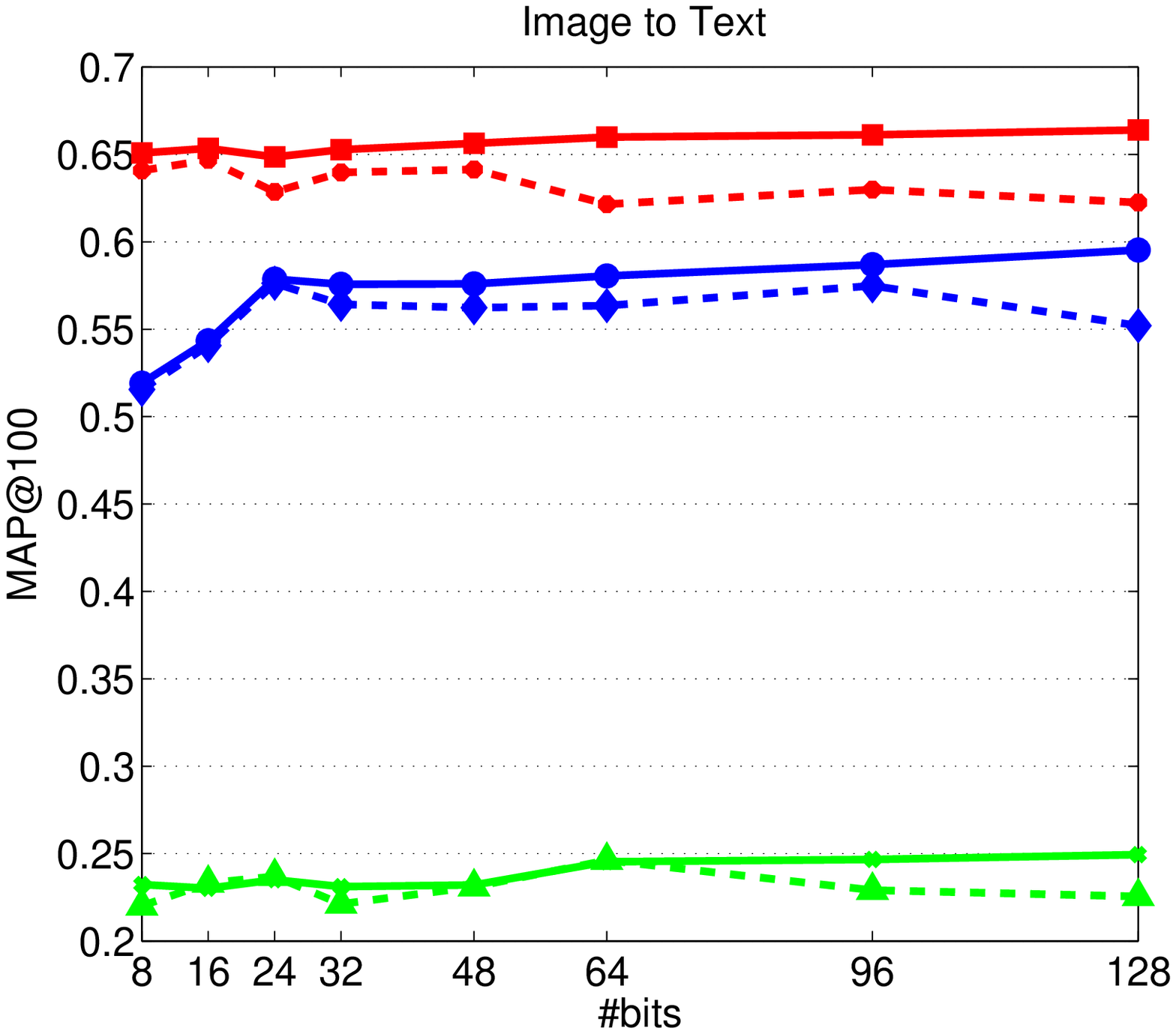}~~
\includegraphics[width=.46\linewidth, clip]{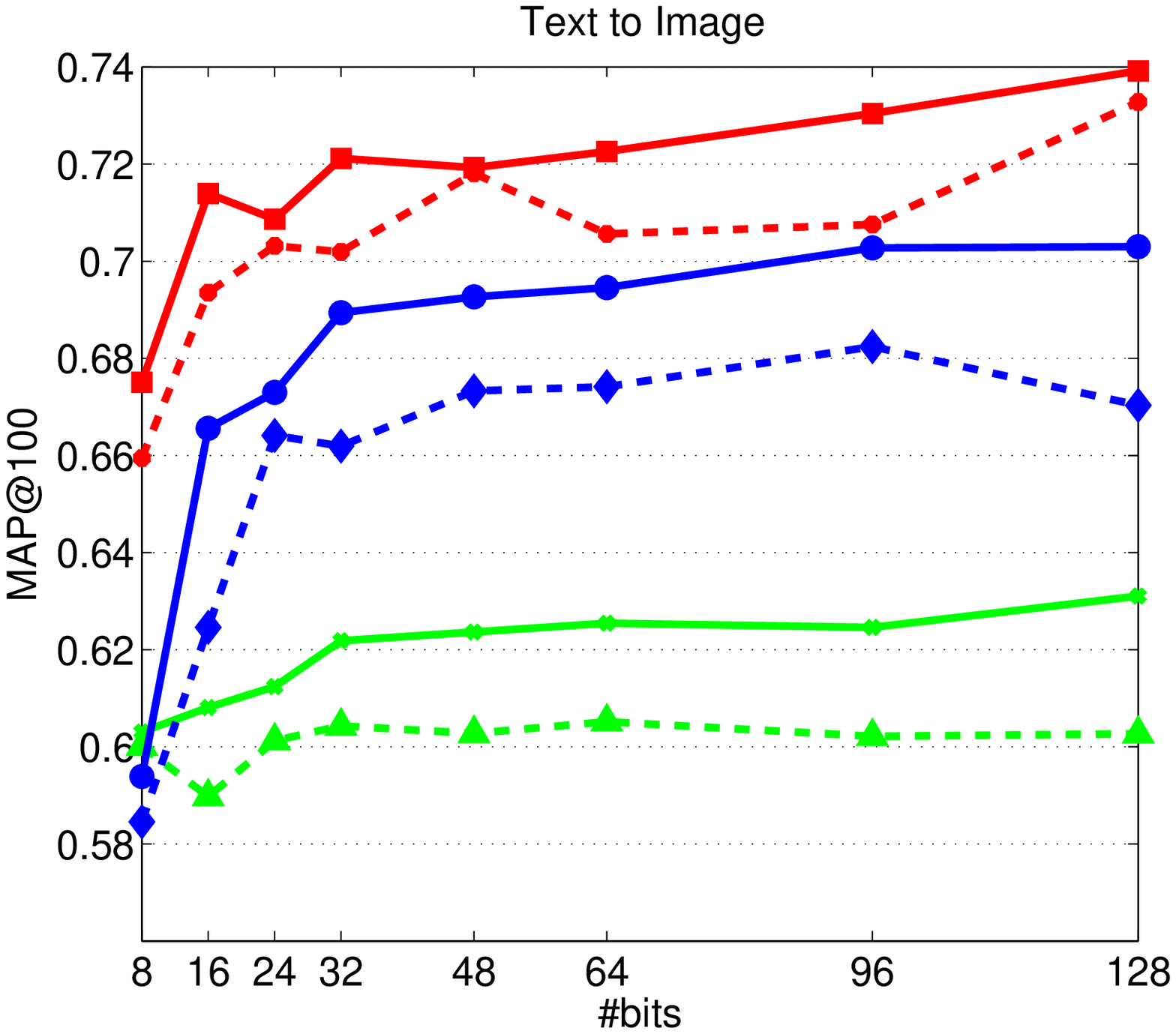}
\caption{Illustrating the effect of the dictionary. The MAP is compared
between CMCQ
and CMCQ ($\bf{C}=\bf{D}$) (using one dictionary for both modalities) on the three datasets.}
\label{fig:samedictionary}
\vspace{-0.5cm}
\end{figure}

\noindent \textbf{The effect of dictionary.}
One possible way for our approach to better catch the intra-document correlation
is to use the same dictionary to quantize both modalities,
i.e., adding constraint $\mathbf{C} = \mathbf{D}$ in the Formulation~\ref{eqn:quantization},
which is
similar to~\cite{LongWY15}.
This might introduce
a closer connection between a pair of image and text,
and hence improve the search quality.
However, our experiments shown in Figure~\ref{fig:samedictionary}
suggest that this is not the case.
The reason might be that
using one dictionary for two modalities
in fact reduces the approximation ability of quantization
when using two dictionaries.

\begin{figure*}[t]
\centering
~~~~~~~~\includegraphics[width = .80\linewidth, clip]{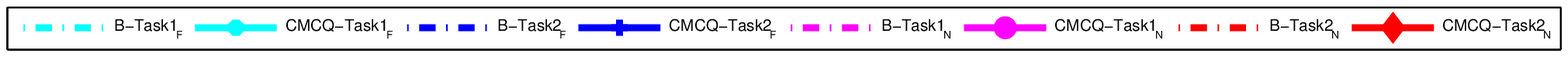}~~\\
(a)\includegraphics[width=.20\linewidth, clip]{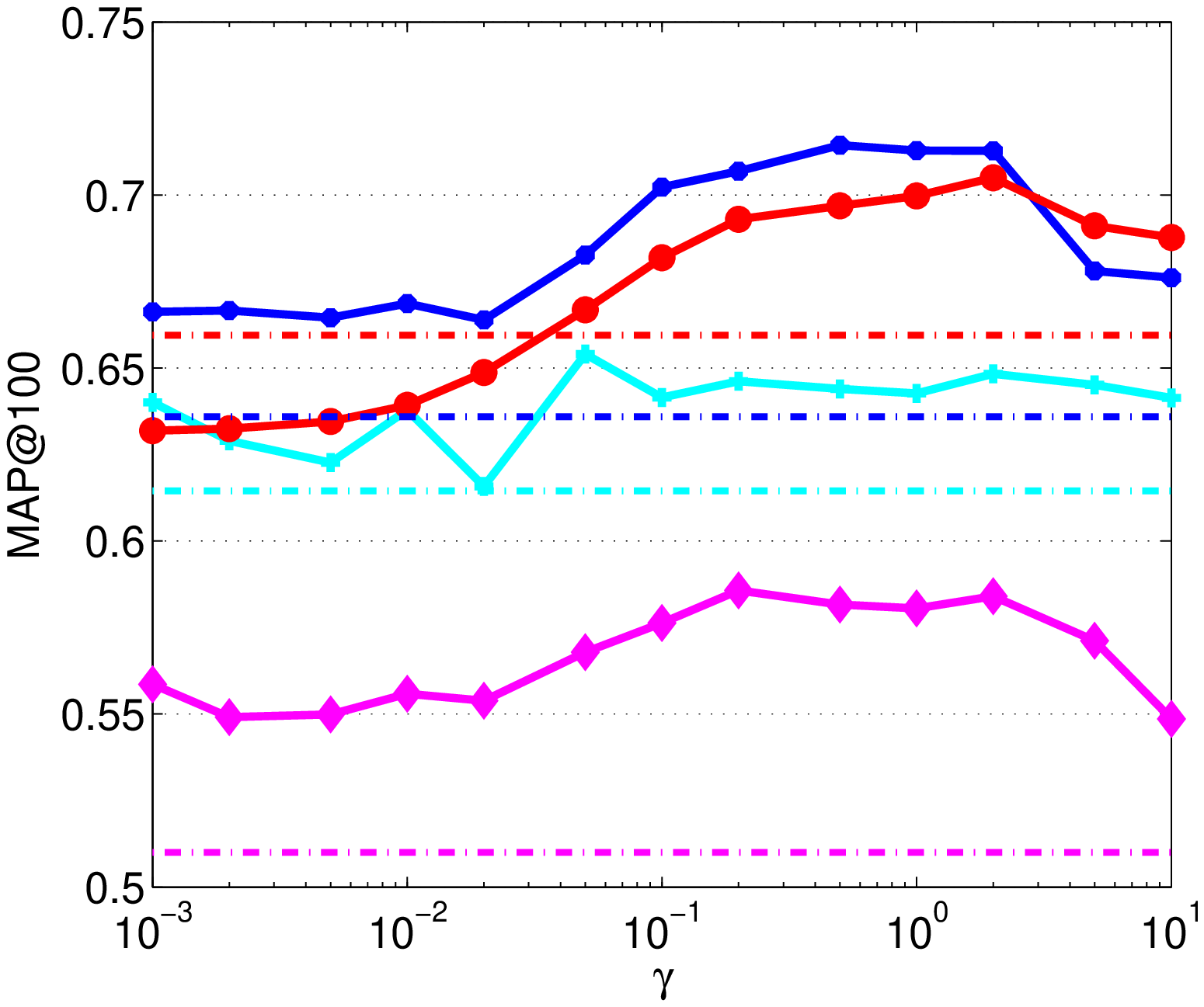}~
(b)\includegraphics[width=.20\linewidth, clip]{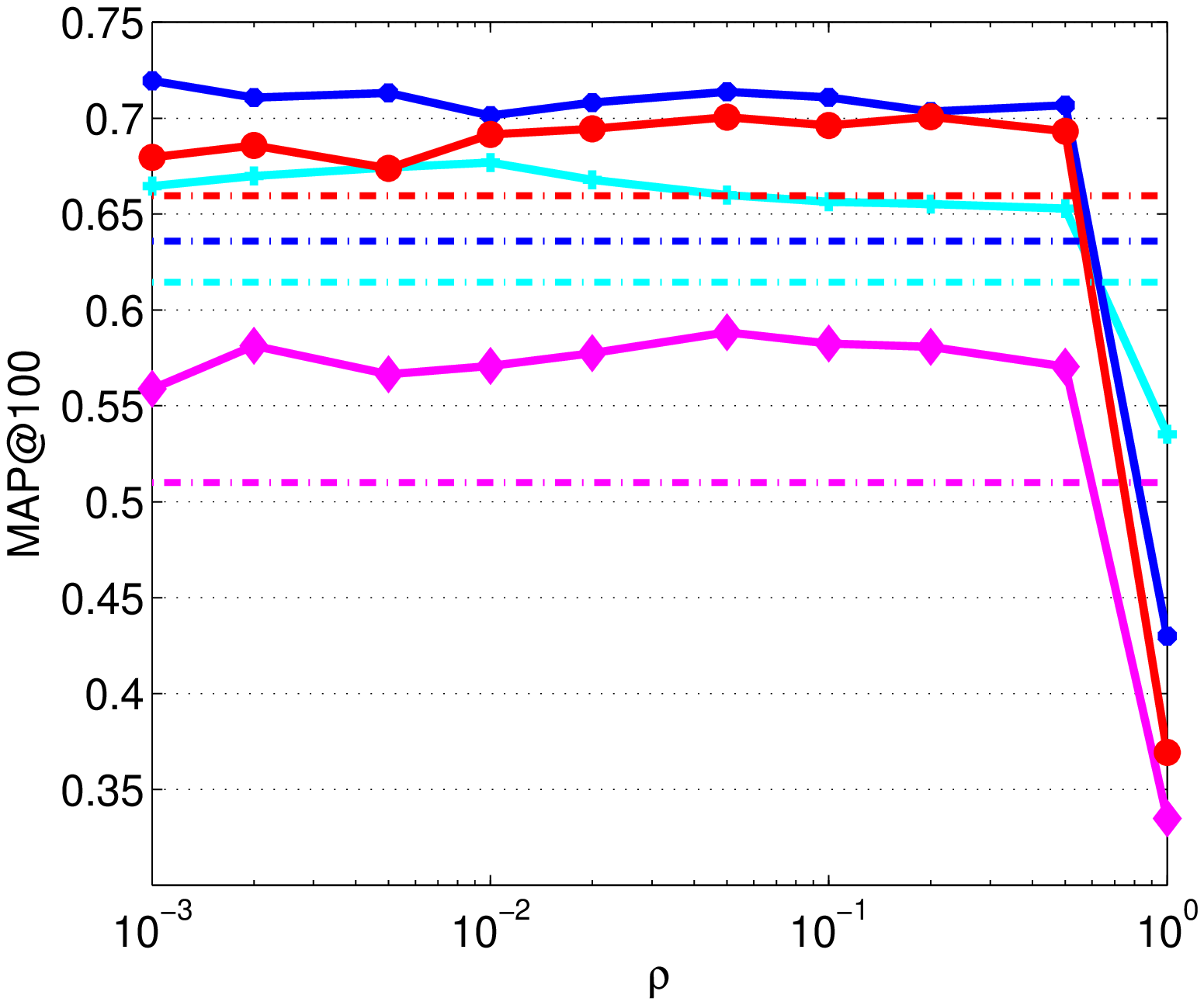}~
(c)\includegraphics[width=.20\linewidth, clip]{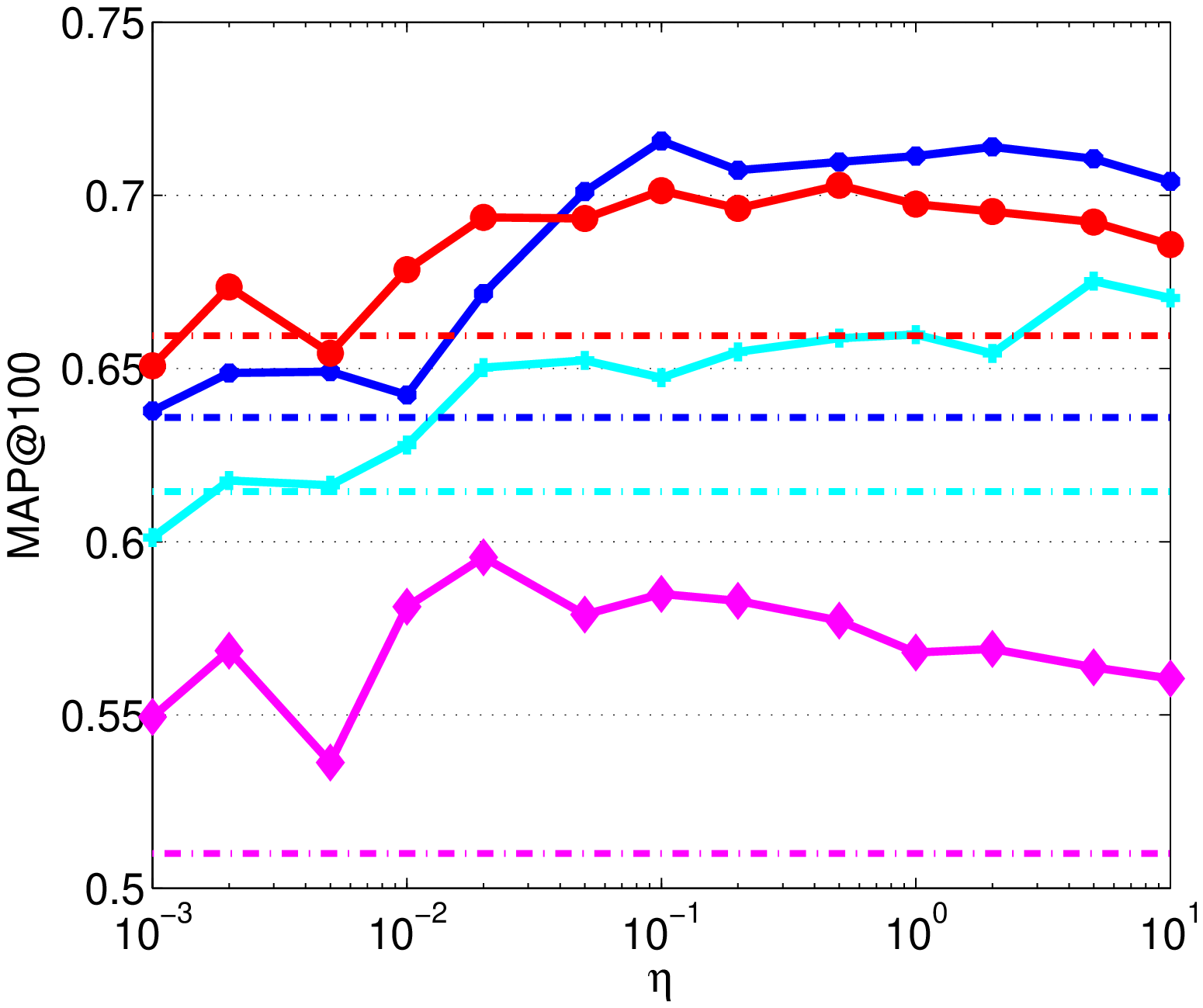}~
(d)\includegraphics[width=.20\linewidth, clip]{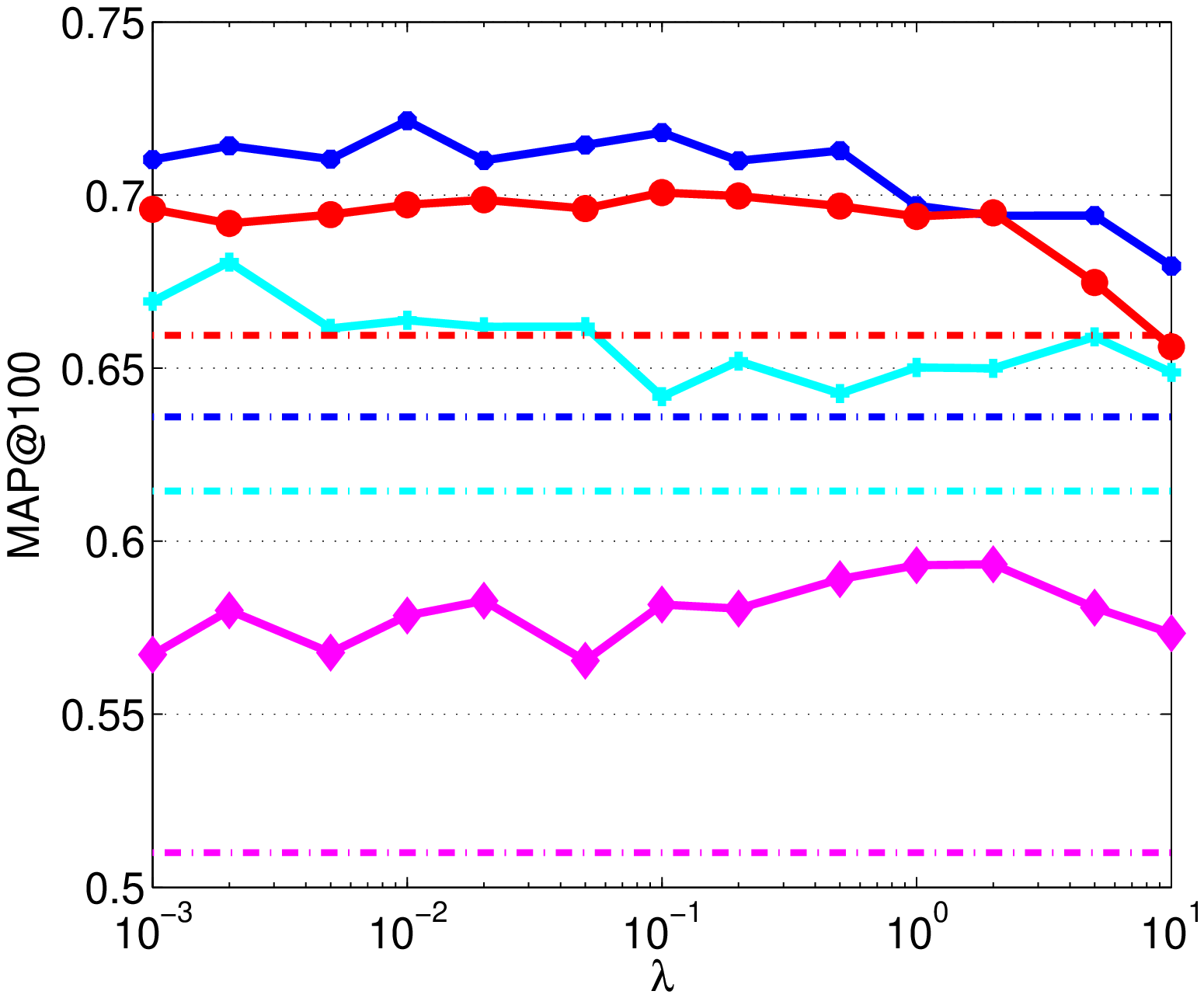}
\caption{Parameter sensitive analysis of our algorithm with respect to (a) $\gamma$, (b) $\rho$, (c) $\eta$, and (d) $\lambda$
over image to text (task1) and
text to image (task2)
on two datasets:
FLICKR$25K$ (F) and NUS-WIDE (N) with 32 bits.
The dashdot line shows the best results obtained by other baseline methods and is denoted as $\text{B}$, e.g., $\text{B-Task1}_\text{F}$ denotes the best baseline results 
the image to text task on FLICKR$25K$.}
\label{fig:parameter}
\vspace{-0.5cm}
\end{figure*}

\noindent \textbf{Parameter sensitive analysis.}
We also conduct the parameter sensitive analysis
to show that our approach is robust to the change of parameters.
The experiments are conducted on FLICKR$25K$
and NUS-WIDE using a validation set,
to form which we randomly sample a subset of the training dataset.
The size of the validation set is 1000 and 2000 respectively
for FLICKR$25K$
and NUW-WIDE.
To evaluate the sensitive of the parameter,
we vary one parameter from 0.001 to 10 (1 for $\rho$)
while keep others fixed.

The empirical results
on the two search tasks (task1: image to text and task2: text to image) are presented in Figure~\ref{fig:parameter}.
It can be seen from the figure that our approach can achieve
superior performance under a wide range of the parameter
values.
We notice that when the parameter $\rho$ gets close to 1, the performance drops suddenly.
The reason might be that with a larger sparsity degree value $\rho$, the learnt image representation in the common space would carry little information since the learnt
$\mathbf{S}$ is a very sparse matrix.


\section{Conclusion}

In this paper, we present a quantization-based compact coding approach,
collaborative quantization,
for cross-modal similarity search.
The superiority of the proposed approach
stems from that it
learns the quantizers for both modalities jointly by
aligning the quantized approximations for
each pair of image and text
in the common space,
which is simultaneously learnt with the quantization.
Empirical results on three multi-modal datasets indicate that the proposed approach
outperforms existing methods.

{\small
\bibliographystyle{ieee}
\bibliography{multimodal}
}

\end{document}